\pdfoutput=1
\documentclass{article}

\usepackage[utf8]{inputenc}
\usepackage{amsmath}        % load before newtxmath

\PassOptionsToPackage{numbers,compress}{natbib}
\usepackage{natbib}

\usepackage[T1]{fontenc}
\usepackage{newtxtext,newtxmath}

\usepackage{nicefrac}
\usepackage{microtype}
\usepackage{xcolor}
\usepackage{graphicx}
\usepackage{booktabs}
\usepackage{multirow}
\usepackage{makecell}
\usepackage{array}
\usepackage{paralist}
\usepackage{enumitem}
\usepackage[font=small,labelfont=bf]{caption}
\usepackage[hyphens]{url}
\usepackage{hyperref}

\hypersetup{
  colorlinks=true,
  linkcolor=blue!45!black,
  citecolor=blue!45!black,
  urlcolor=blue!55!black,
  breaklinks=true
}

\newcolumntype{P}[1]{>{\centering\arraybackslash}p{#1}}

\title{Beyond Memory Leaderboards: Evaluating Scientific Memory as Budgeted Context Restoration}

\author{%
Maksim Sheverev\\
\textit{Quantellence Research}\\
\texttt{}
\and
David Finkelstein\\
\textit{Quantellence Research}\\
\texttt{}
\and
  Sergey Nikolenko\\
  \textit{St. Petersburg Department of the}\\
  \textit{Steklov Institute of Mathematics, St. Petersburg, Russia}\\
  \textit{St. Petersburg State University, St. Petersburg, Russia}\\
  \texttt{sergey@logic.pdmi.ras.ru}
}
\date{\today}

\begin{document}

\maketitle

% =============================================================================
\begin{abstract}
Long-term memory is becoming a core component of LLM agents, but most memory
benchmarks evaluate conversations or compact summaries, while research agents
need to restore evidence from full scientific papers. We introduce two
full-text scientific-memory benchmarks, \textsc{Public AI Memory}
(\textsc{Paim}; 81 papers, 66 audited questions) and \textsc{Public
Transformers} (\textsc{PTr}; 252 papers, 98 audited questions). We evaluate
eight memory/retrieval systems, including our own proposed system
\emph{Theoria}, plus a no-retrieval baseline under a shared
context-restoration protocol. Our results show that memory leaderboards are
not interpretable without the full protocol: ingestion granularity, raw-text
preservation, retrieval budget, retrieval modality, rubric
audit, and judge choice all affect the outcome. For example, on \textsc{Paim}
Graphiti wins convincingly but uses 2.6M characters of retrieved context per
query, and after controlling for retrieval budget the lead disappears. On
\textsc{PTr}, for the systems where BM25 retrieval can be added cleanly, the
sparse--dense hybrid is the single most significant intervention: hybrid
variants of Simple RAG, Mem0, and Theoria tie for the lead within 0.03 points.
Multi-judge and human side-by-side calibration show that LLM-as-a-judge
rankings are consistent across frontier judges and agree with human evaluation,
with an effective resolution of roughly one point on a ten-point scale. We
argue that scientific memory should be evaluated as \emph{budgeted,
modality-aware context restoration} rather than as an unconstrained
architecture leaderboard, and we release the datasets, harness, raw outputs,
judgments, and scripts to reproduce our results and serve as tools for such evaluation.
Our code is available at
\url{https://gitlab.com/quantellence/research/scientific-recall-bench}, and the
datasets are available at
\url{https://huggingface.co/datasets/quantellence/srb-data}.
\end{abstract}

% =============================================================================
\section{Introduction}
\label{sec:intro}
% =============================================================================

LLM agents increasingly rely on long-term memory mechanisms: vector stores,
knowledge graphs, episodic logs, extracted facts, and hybrid retrieval
pipelines. The current approaches to evaluating agent memory are largely
inherited from conversational settings, with multi-session dialogues, user
preferences, and fact updating; they usually treat memory as a
\emph{store-and-recall} module to be ranked on benchmarks such as
LongMemEval~\citep{wu2025longmemeval}, LoCoMo~\citep{maharana2024locomo}, or
BEAM~\citep{tavakoli2026beam}.

\paragraph{Why memory at all?}
Before discussing how to evaluate memory, let us recall why agents need
an explicit memory subsystem when context windows already reach one or two
million tokens and retrieval-augmented generation (RAG) can fill them. There
are at least four reasons, all of which recur throughout the systems we survey
in Section~\ref{sec:related}.
\begin{enumerate}[label=(\arabic*)]
  \item \emph{Long contexts degrade.} The \emph{lost-in-the-middle}
  effect~\citep{liu2024lostmiddle} shows that models
attend well to the beginning and end of a long context but lose information in
the middle, and interference grows with context length; a million-token window
may be a poor substitute for curated memory. 
  \item \emph{Long contexts are expensive}, both in money and in latency, because re-reading the entire
history on every turn scales super-linearly with input size. 
  \item \emph{There is a benchmark-to-deployment gap.} Systems that score near-perfectly on
conversational recall benchmarks can drop to 40--60\% when the same facts must
be \emph{used} inside a downstream task such as web navigation or constrained
planning~\citep{he2026memoryarena}; passive recall and active use are not the
same skill. 
  \item \emph{Memory must update.} If an agent learned months ago that
a user lives in Berlin and the user has since moved to Lisbon, the system has to
decide whether to forget, overwrite, or version the stale fact. A useful memory
is therefore not a log or a database but a system that answers a family of
questions: what to store, how to store it, how to update it, what to return at
query time, how to fuse evidence, how to flag contradictions, and when to
forget.
\end{enumerate}

\paragraph{Scientific memory is context restoration.}
In our opinion, the standard conversational framing does not transfer cleanly
to \emph{research} agents. When an agent reads a research paper, the relevant
unit of memory is rarely a single fact: it is a conditional claim, the evidence
behind it, the regimes in which it holds, the methods used to obtain it, and
how it relates to other papers. The agent's task is not to remember a user
preference but to reconstruct, on demand, an interconnected body of literature
relevant to a research question. We call this task \emph{context restoration}:
the memory must return enough grounded evidence for a downstream model to
compose a faithful answer. A scientific memory system is therefore valuable
not only because it stores facts but because it can restore the right context
for a downstream reasoning process, at the right granularity and within a
realistic budget.

In this work we present two new full-text question answering (QA) benchmarks on
research papers and show experimentally that the distinction between recall and
context restoration has empirical consequences large enough to invert standard
memory leaderboards in some settings. Our datasets consist of recent research
papers on a given topic together with questions of varying difficulty that
require either precise search or broad synthesis across several papers; we
release \textsc{Public AI Memory} (\textsc{Paim}) on AI agent memory and
\textsc{Public Transformers} (\textsc{PTr}) on recent Transformer
architectures. We also present \emph{Theoria}, our own approach to memory for
AI research agents, which is competitive with the strongest baselines under
matched retrieval and wins on specific difficulty tiers.

\begin{figure}[!t]
  \centering
  \includegraphics[width=\linewidth]{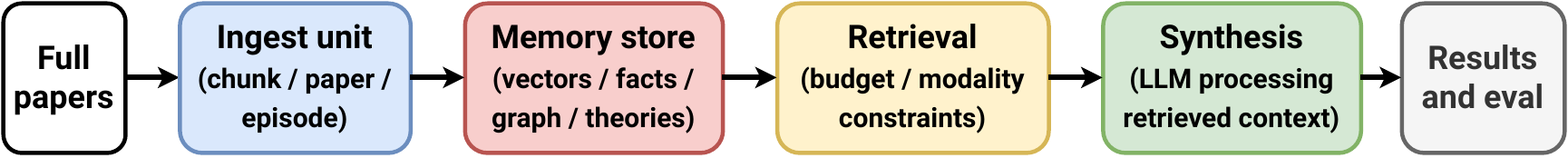}
  \caption{The context-restoration protocol. A memory system can differ at
  every stage: how full papers are split into ingestion units, how those units
  are represented in the store, how much and what kind of context is returned
  at retrieval time (the budget $B$ and modality $M$), and whether the final
  answer is synthesized internally by the system or externally by a shared
  model.}
   % We hold the synthesis model (gpt-4.1-mini, $T{=}0$) and the primary
  % judge (Gemini~3.1~Pro) fixed, and vary the retrieval budget and modality
  % explicitly while reporting rubric audit and judge choice as
  % evaluation-hygiene controls.}
  \label{fig:pipeline}
\end{figure}

Figure~\ref{fig:pipeline} shows the evaluation protocol. We focus on the
\emph{retrieval budget} ($B$, characters returned per query) and the
\emph{retrieval modality} ($M$, dense vs.\ dense$+$lexical), and we compare
memory approaches under the \emph{same} synthesis model. Our evaluation yields three main
findings.

{First, comparisons can be dominated by the sheer volume of retrieved
context.} Graphiti~\citep{graphiti} wins on \textsc{Paim} largely because it
returns the raw text of matched episodes, up to 2.6M characters per
query, and when the retrieval budget is constrained it drops to the bottom of
the leaderboard. In a fair comparison with fixed budget $B\approx 30$K characters, Simple
RAG ($7.25$) $\approx$ Theoria ($7.19$) $>$ Mem0 ($6.97$) $>$ Hindsight
($6.53$) $>$ Cognee ($5.28$) $\approx$ Graphiti's knowledge-graph output
($5.27$), so the original win was entirely due to context volume.

{Second, corpus structure and retrieval modality matter.} On \textsc{Paim}
we see a dense top cluster with Simple RAG, Theoria, and Mem0, whereas on
\textsc{PTr}, whose named entities (model names, kernel names, hyperparameters)
are lexically distinctive, structured memories (Theoria, Mem0) pull ahead of
Simple RAG under dense retrieval. Interestingly, adding lexical BM25 retrieval
to the dense embeddings and thus producing a sparse--dense hybrid was the single largest
intervention we measured, significantly improving the scores for many memory systems.
% : on \textsc{PTr} at $B\approx 50$K, Mem0$+$BM25
% ($9.24$), Simple~RAG$+$BM25 ($9.22$), and Theoria$+$BM25 ($9.21$) tie within
% $0.03$ points, each about $0.3$ above its dense-only counterpart.

{Third, we find LLM-as-a-judge evaluation to be quite reliable, especially on rank.}
A subset re-evaluation across three judges (Gemini~3.1~Pro, Sonnet~4.6,
GPT-5.4) on \textsc{Paim} gives rank Spearman $\rho \in [0.90, 0.97]$;
on \textsc{PTr}, DeepSeek~V4~Pro rankings correlate with Gemini at Pearson
$r=0.93$ over $n=946$ valid paired cells even though DeepSeek is systematically
more critical. A human study with $112$ blinded side-by-side (SxS) votes shows
high agreement with the LLM judge, scaling cleanly with the score gap.

\paragraph{Contributions.}
\begin{inparaenum}[(1)]
\item Two full-text scientific-memory benchmarks with audited rubrics,
  \textsc{Paim} and \textsc{PTr}.
\item A unified experimental harness covering eight memory/retrieval systems and
  a no-memory baseline, with hybrid BM25 variants for three of them and a
  budget-and-modality retrieval protocol.
\item Multi-judge LLM calibration and $112$ blinded human SxS votes that
  establish the resolution of the LLM judge ($\sim$$1$ point on a $0$--$10$
  scale) and underwrite our experimental claims.
\item \emph{Theoria}, our own three-layer scientific-memory system, with an
  ablation (\emph{Prism}) that isolates its community/theory layers.
\end{inparaenum}
We release the full datasets, experimental results, and scripts so that every
figure and table in this paper can be reproduced: the \texttt{scibench}
evaluation harness, per-system adapters, raw outputs, judgments, and analysis
scripts are available on
GitLab,\footnote{\url{https://gitlab.com/quantellence/research/scientific-recall-bench}}
and both benchmark corpora with their audited rubrics are hosted on Hugging
Face.\footnote{\url{https://huggingface.co/datasets/quantellence/srb-data}}

The remainder of the paper is organized as follows.
Section~\ref{sec:related} surveys agent-memory systems and benchmarks and
positions our work; Section~\ref{sec:systems} describes the eight evaluated
systems; Section~\ref{sec:bench} introduces the two benchmarks and the rubric
audit; and Section~\ref{sec:theoria} presents \emph{Theoria}.
Section~\ref{sec:eval} presents our experimental evaluation, including the setup,
\textsc{Paim} and \textsc{PTr} budget-and-modality sweeps, statistical significance,
and judge calibration, while Section~\ref{sec:qualitative} analyzes representative answers
qualitatively (including how a single system's answer shifts with the retrieval
budget). Section~\ref{sec:discussion} discusses implications and limitations
and draws lessons for the evaluation infrastructure, and
Section~\ref{sec:conclusion} concludes the paper.
Appendix~\ref{app:examples} collects several more sample questions and answers.

% =============================================================================
\section{Related work}
\label{sec:related}
% =============================================================================

Agent memory has grown into a large and fast-moving field; a useful heuristic is
that when a problem has twenty competing solutions, none of them works
decisively well yet. This section surveys the landscape we benchmark against. We
organize it by idea rather than chronology, and then we discuss in Section~\ref{sec:systems}
the specific systems that we evaluate in this work. For broader treatments we refer to four
recent surveys, each taking a different cut.

\subsection{Surveys and taxonomies}
\label{sec:surveys}

There are three main recent surveys of memory for LLM-based agents.
\emph{Memory in the Age of AI Agents}~\citep{hu2025memorysurvey} organizes
memory along three orthogonal axes---\emph{forms} (token-level / parametric /
latent), \emph{functions} (factual / experiential / working), and
\emph{dynamics} (formation / evolution / retrieval)---and adds dedicated
sections on benchmarks, open-source frameworks, and frontier directions.
\emph{CoALA}~\citep{sumers2024coala} ports Tulving's
working/episodic/semantic/procedural split from cognitive psychology to LLM
agents; this approach has become a default reference taxonomy.
\citet{jiang2026anatomy} survey eleven architectural patterns and document
\emph{benchmark saturation} and evaluation fragility, that is, sensitivity to the judge
model and to the backbone LLM. They explain why reported numbers are often
inflated even without any intent to mislead. In addition, \citet{zhang2024memorysurvey}
provide an earlier systematic review that is useful mainly for the history of
the problem.

It is helpful to keep these three complementary taxonomies in mind. The
\emph{neuropsychological} taxonomy (CoALA) classifies memory by what it stores:
working memory (what is in context now), episodic memory (timestamped past
events), semantic memory (facts about the world), and procedural memory (learned
skills and behaviors). The \emph{operational} taxonomy classifies memory by the
atomic operations a system must support; almost all modern systems implement
three, called \emph{formation / management / retrieval} in academic surveys and
\emph{retain / recall / reflect} in the Hindsight framing. The surveys agree
that retrieval is comparatively well understood, both via classical information 
retrieval and with the extensively developed RAG tools, while the hardest, 
least-solved problems remain in management: conflict resolution, temporal reasoning, selective
forgetting, and knowledge updating. Finally, the \emph{methodological} taxonomy
classifies systems by their dominant mechanism, which is the organization we use
below: store-first vector search, extract-and-update, OS-like hierarchies,
temporal knowledge graphs, self-editing notes, verbal reinforcement learning,
RL-trained memory management, multi-strategy parallel retrieval with fusion,
graph-based associative recall, and bio-inspired consolidation.

\subsection{Three foundational systems}
\label{sec:foundations}

\emph{Generative Agents}~\citep{park2023generative} introduced the
\emph{memory stream}: an append-only log of natural-language observations, each
scored at retrieval time by a linear combination of three normalized signals,
$$\text{score} = \alpha_{\text{rec}}\,\text{recency} +
\alpha_{\text{imp}}\,\text{importance} + \alpha_{\text{rel}}\,\text{relevance},$$
where recency is an exponential decay, importance is an LLM-assigned $1$--$10$
``poignancy'' rating, and relevance is embedding cosine similarity. In the
released implementation all three weights are simply $1$, yet the scheme works
well enough that most retrieval-based memories still use it. Generative
Agents also introduced \emph{reflection}: periodically the agent poses
high-level questions about recent experience, retrieves relevant memories, and
synthesizes higher-level insights that are written back into the stream, growing
a tree where meaning is supposed to crystallize out of raw observations. Almost everything
since is either a refinement of one component of this formula or a replacement
of one component by something more sophisticated.

\emph{MemGPT}~\citep{packer2023memgpt}, now productionized as
\emph{Letta}~\citep{letta}, treats memory as an operating system: a fixed-size main
context (system instructions, a read/write working block, and a FIFO queue with
recursive summarization of the evicted tail) backed by external recall and
archival stores, with explicit tool calls
(\texttt{core\_memory\_append}, \texttt{archival\_memory\_search}, \dots) and
\emph{memory-pressure warnings} analogous to page faults. Its lasting
contribution is the principle that \emph{the agent manages its own memory},
which has since been reproduced in many different forms, including the
\texttt{CLAUDE.md}-style memory files used by modern coding agents.

\emph{Reflexion}~\citep{shinn2023reflexion} replaces parameter updates with
verbal self-critique: an actor produces a trajectory, an evaluator scores it,
and a self-reflection module writes a natural-language post-mortem that is
appended to the next attempt's prompt. The authors describe the text feedback as
a \emph{semantic gradient}, a specific improvement direction expressed in
tokens rather than numbers, and observe, somewhat counter-intuitively, that more
reflections are not better: the buffer is capped at $1$--$3$ entries because
larger buffers drown the agent in contradictory advice. Compression of memory is
often more important than its accumulation, and this is a theme we will return to.

\begin{table}[!t]
  \centering
  \caption{The surveyed agent-memory systems ranked across three axes; $\bullet$ marks a primary, defining
  property, $\circ$ a secondary one.}
  % \emph{Theoria} (last row) is our own system
  % (Section~\ref{sec:theoria}); its ablation \emph{Prism} and the retrieval
  % baselines are characterized in Table~\ref{tab:systems}.}
  \label{tab:survey}
  \scriptsize
  \setlength{\tabcolsep}{2.6pt}
  \renewcommand{\arraystretch}{1.06}
  \begin{tabular}{@{}>{\raggedright\arraybackslash}p{1.4cm} >{\centering\arraybackslash}p{.6cm} >{\raggedright\arraybackslash}p{1.80cm} *{4}{>{\centering\arraybackslash}m{0.10cm}} *{3}{>{\centering\arraybackslash}m{0.15cm}} >{\raggedright\arraybackslash}p{\dimexpr\linewidth-4.70cm-\tabcolsep*20\relax}@{}}
    \toprule
    & & & \multicolumn{4}{c}{\textbf{CoALA}} & \multicolumn{3}{c}{\textbf{Oper.}} & \\
    \cmidrule(lr){4-7}\cmidrule(lr){8-10}
    \textbf{System} & \textbf{Year} & \textbf{Mechanism} &
    \textbf{W} & \textbf{E} & \textbf{S} & \textbf{P} &
    \textbf{Rt} & \textbf{Rc} & \textbf{Rf} & \textbf{Key idea / signature} \\
    \midrule
    \multicolumn{11}{@{}l}{\textit{Memory streams, OS hierarchies, and verbal reinforcement}}\\
    \addlinespace[1pt]
    Generative Agents \citep{park2023generative} & 2023 & Weighted memory stream &
      & $\bullet$ & $\circ$ & & & $\bullet$ & $\bullet$ &
      Recency--importance--relevance scoring of an observation stream; reflection synthesizes higher-level insights \\
    MemGPT / Letta \citep{packer2023memgpt,letta} & 2023 & Tiered OS memory $+$ paging &
      $\bullet$ & $\circ$ & $\circ$ & & $\bullet$ & $\circ$ & $\circ$ &
      Memory as a virtual OS; the agent self-manages tiers via tool calls, with memory-pressure ``page faults'' \\
    Reflexion \citep{shinn2023reflexion} & 2023 & Verbal-RL critique buffer &
      & $\bullet$ & & $\bullet$ & & $\circ$ & $\bullet$ &
      Verbal self-critique as a ``semantic gradient'' appended to the next attempt; capped $1$--$3$-entry buffer \\
      \midrule
    % \addlinespace[2pt]
    \multicolumn{11}{@{}l}{\textit{Graph-based and temporal memory}}\\
    \addlinespace[1pt]
    HippoRAG 1/2 \citep{gutierrez2024hipporag,gutierrez2025hipporag2} & 2024 / 2025 & KG $+$ PPR &
      & & $\bullet$ & & $\circ$ & $\bullet$ & &
      One Personalized-PageRank pass over a schemaless KG replaces multi-hop LLM traversal; v2 fixes the factoid regression \\
    Zep / Graphiti \citep{graphiti} & 2025 & Bi-temporal knowledge graph &
      & $\circ$ & $\bullet$ & & $\bullet$ & $\bullet$ & $\circ$ &
      Facts carry \texttt{valid\_at}/\texttt{invalid\_at} timestamps; embedding $+$ BM25 $+$ graph search with async indexing \\
    MAGMA \citep{jiang2026magma} & 2026 & Multi-graph $+$ temporal engine &
      & $\bullet$ & $\bullet$ & & $\bullet$ & $\bullet$ & &
      Multi-relational entity/episodic/temporal/semantic subgraphs plus a temporal inference engine that normalizes time \\
    \midrule
    % \addlinespace[2pt]
    \multicolumn{11}{@{}l}{\textit{Mutable, learned, and bio-inspired memory}}\\
    \addlinespace[1pt]
    A-MEM \citep{xu2025amem} & 2025 & Self-editing Zettelkasten notes &
      & $\circ$ & $\bullet$ & & $\circ$ & $\circ$ & $\bullet$ &
      Atomic linked notes; adding a note rewrites the descriptions and tags of the old notes it links to (``memory evolution'') \\
    Memory-R1 \citep{yan2025memoryr1} & 2025 & RL-trained policy &
      & & $\bullet$ & & $\bullet$ & & $\bullet$ &
      An RL policy learns ADD/UPDATE/DELETE/NOOP from downstream answer reward; only $\sim$$152$ training pairs needed \\
    LightMem \citep{lightmem} & 2025 & Sleep-time consolidation &
      $\bullet$ & & $\circ$ & & $\circ$ & & $\bullet$ &
      Atkinson--Shiffrin stages; consolidation moved off the inference path into a background ``sleep'' phase \\
    SleepGate \citep{sleepgate} & 2026 & KV-cache consolidation &
      $\bullet$ & & & & & & $\bullet$ &
      Sleep-inspired consolidation applied to the KV-cache itself, targeting proactive interference from stale memories \\
    EverMemOS \citep{evermemos} & 2026 & MemCell / MemScene store &
      & $\bullet$ & $\bullet$ & & $\bullet$ & $\bullet$ & $\circ$ &
      Heterogeneous items become \emph{MemCells} aggregated into \emph{MemScenes} for structured long-horizon retrieval \\
    Nemori \citep{nemori} & 2025 & Surprise-gated admission &
      & $\bullet$ & $\circ$ & & $\bullet$ & & $\circ$ &
      A free-energy / predictive-coding rule admits an episode only when it is a ``surprise'' vs.\ current knowledge \\
    \midrule
    % \addlinespace[2pt]
    \multicolumn{11}{@{}l}{\textit{Production frameworks and multi-strategy retrieval}}\\
    \addlinespace[1pt]
    Mem0 \citep{mem0} & 2025 & Extract-and-update facts &
      & & $\bullet$ & & $\bullet$ & $\bullet$ & $\circ$ &
      One-line \texttt{add()} extracts atomic facts, deduplicates, and ADD/UPDATEs on conflict; vector $+$ optional graph/KV \\
    Cognee \citep{cognee} & 2024 & ECL typed-KG completion &
      & & $\bullet$ & & $\bullet$ & $\bullet$ & &
      An Extract--Cognify--Load pipeline builds a typed-schema KG and answers via internal graph completion \\
    MemPalace \citep{mempalace} & 2026 & Spatial hierarchy, $0$-LLM write &
      & $\circ$ & $\bullet$ & & $\bullet$ & $\circ$ & &
      Wings/rooms/halls ``palace'' with a deterministic zero-LLM write path; headline recall was a vector-store confound \\
    Hindsight \citep{hindsight} & 2025 & 4 networks $+$ multi-strategy RRF &
      & $\bullet$ & $\bullet$ & & $\circ$ & $\bullet$ & $\bullet$ &
      Four epistemic networks (world/experience/observation/opinion); TEMPR fuses semantic\,$+$\,BM25\,$+$\,graph\,$+$\,time by RRF; dispositions \\
    Framework memories$^{\dagger}$ & 2023--2025 & Built-in framework memory &
      $\circ$ & & $\circ$ & $\bullet$ & $\circ$ & $\bullet$ & $\circ$ &
      LangMem (procedural self-prompt rewriting); LlamaIndex (composable memory blocks $+$ token budget); CrewAI (shared weighted memory) \\
    \midrule
    % \addlinespace[2pt]
    \multicolumn{11}{@{}l}{\textit{Structured scientific memory (this work)}}\\
    \addlinespace[1pt]
    Theoria \textit{(this work)} & 2026 & Evidence $+$ community $+$ theory stack &
      & & $\bullet$ & & $\bullet$ & $\bullet$ & $\circ$ &
      Three-layer evidence~$+$~community~$+$~theory store: multi-aspect claim extraction over a RAPTOR tree, Leiden community-routed retrieval, typed supports/contradicts links, and confidence-rated theory statements \\
    \bottomrule
  \end{tabular}
  % \\[2pt]
  % {\scriptsize $^{\dagger}$Agent-framework memory subsystems discussed in
  % Section~\ref{sec:frameworks}; cited there by URL rather than as papers.}
\end{table}

Table~\ref{tab:survey} gives a general overview of influential memory systems for AI agents that we detail below.
It ranks systems by three axes: rows are
  grouped by \emph{dominant mechanism} (methodological taxonomy),
  {CoALA memory type} columns are the neuropsychological taxonomy
  (\textbf{W}orking / \textbf{E}pisodic / \textbf{S}emantic / \textbf{P}rocedural),
  and operation emphasis columns represent the operational taxonomy
  (\textbf{Rt}\,$=$\,retain/formation, \textbf{Rc}\,$=$\,recall/retrieval,
  \textbf{Rf}\,$=$\,reflect/management).

\subsection{Graphs and neuroscience}
\label{sec:graphs}

A parallel line of work treats memory as a graph. 
\emph{HippoRAG}~\citep{gutierrez2024hipporag}
maps neocortex / parahippocampus / hippocampus onto an LLM doing open
information extraction, a retrieval encoder that adds synonymy edges above a
similarity threshold, and a schemaless knowledge graph; at query time it runs
{Personalized PageRank} seeded by query entities instead of multi-hop LLM
traversal, which is both cheaper and more accurate than one-shot RAG.
\emph{HippoRAG~2}~\citep{gutierrez2025hipporag2} fixes the first version's
weakness on simple factoid questions (where graph diffusion blurred easy cases)
with a more careful integration of retrieved passages. The lesson is that such
systems work exactly as well as their open information extraction does: many
errors now come from entity extraction rather than from the graph search itself.

\emph{Zep / Graphiti}~\citep{graphiti} brings this idea to production with a
\emph{bi-temporal knowledge graph}: every fact stores both $t_{\text{valid}}$
(when it became true) and $t_{\text{invalid}}$ (when it ceased to be true), so a
``where did the user live in February?'' query returns the right answer even
after the user moves. Graphiti combines embedding search, BM25, and graph
traversal, with asynchronous background extraction and indexing; it is one of
the most developed graph memories for temporal
recall. \emph{MAGMA}~\citep{jiang2026magma} couples multi-relational subgraphs
(entity / episodic / temporal / semantic) with a temporal inference engine that
normalizes temporal expressions into a chronological representation.

\subsection{Mutable, learned, and bio-inspired memory}
\label{sec:mutable}

Most of the systems above only \emph{add} elements, but human memory is
continually re-consolidated. \emph{A-MEM}~\citep{xu2025amem} casts each memory
as a Zettelkasten note (raw content, timestamp, LLM-extracted keywords and tags,
a generated contextual description, an embedding, and links); when a new note is
added, the system links it to its nearest neighbors and, crucially, rewrites the
descriptions and tags of the \emph{old} linked notes in light of the new one---a
mechanism the authors call \emph{memory evolution}.

\emph{Memory-R1}~\citep{yan2025memoryr1} takes the further step of
\emph{learning} the memory-management policy: instead of hand-coded heuristics,
an RL-trained agent chooses an operation (ADD/UPDATE/DELETE/NOOP) for each new item, rewarded by
downstream answer correctness (PPO and GRPO converge to similar points). In their
experiments, $\sim$$152$ QA pairs sufficed to beat strong prior systems; this is, in our view,
one of the most promising directions in the area, even if the best
ready-to-use systems today are not yet RL-trained.

Several recent systems borrow explicitly from models of human memory.
\emph{LightMem}~\citep{lightmem} implements Atkinson--Shiffrin-style sleep-time
consolidation, separating background consolidation from inference to speed up
queries. \emph{SleepGate}~\citep{sleepgate} addresses \emph{proactive
interference}---the degradation of retrieval as stale information
accumulates---by updating the KV-cache rather than an abstract store.
\emph{EverMemOS}~\citep{evermemos} introduces \emph{MemCells} that aggregate
heterogeneous items into \emph{MemScenes} for retrieval, and
\emph{Nemori}~\citep{nemori} applies a free-energy / predictive-coding
principle, admitting a new episode into memory only when it is a
\emph{surprise} relative to what the existing knowledge predicts.

\subsection{Open-source frameworks}
\label{sec:frameworks}

In practice most developers use a ready-made framework.
\emph{Mem0}~\citep{mem0} is among the most popular: a one-line \texttt{add()}
extracts facts and preferences with an LLM, deduplicates, and updates on
conflict, while \texttt{search()} ranks by relevance/importance/recency over a
vector store with optional graph and key-value backends; it remains an easy and
strong baseline. \emph{Cognee}~\citep{cognee} ships a standalone
``Extract~$\to$~Cognify~$\to$~Load'' (ECL) pipeline that builds a typed
knowledge graph and answers via graph completion. Agent frameworks bundle their
own memory: LangChain's LangMem implements procedural memory by letting the
agent rewrite its own system prompt,\footnote{\url{https://langchain-ai.github.io/langmem/}}
LlamaIndex offers composable memory blocks with a priority token
budget,\footnote{\url{https://www.llamaindex.ai/}} and CrewAI provides a shared
multi-agent memory with explicit recency/similarity/importance
weights.\footnote{\url{https://github.com/crewAIInc/crewAI}} During 2025 all
three major providers (OpenAI, Anthropic, Google) added persistent,
importable/exportable memory, with philosophically different
defaults---transparent automatic memory versus memory loaded only through
explicit, user-visible tool calls.\footnote{\url{https://code.claude.com/docs/en/memory}}

The recent \emph{MemPalace}~\citep{mempalace} framework is an instructive
cautionary tale for exactly the evaluation problems this paper studies. It
organizes memory as a spatial hierarchy (a ``memory palace'' of wings, rooms,
halls, tunnels, closets, and drawers) with a zero-LLM write path and 
aggressive regex-based compression, which is an elegant, human-interpretable
design. However, independent analysis showed that its headline result
($96.6\%$~Recall@5 on LongMemEval, claimed as the highest published) was
obtained in a mode that stores documents verbatim in a vector database and uses
its {default} embedding search, with none of the ``palace'' machinery in
the retrieval path; enabling the palace structure \emph{lowered} the numbers.
In this work, we aim for a controlled leaderboard that makes all parameters
explicit and can distinguish between the contributions of different components.
% The headline metric was measuring the default vector store, not the proposed
% architecture---precisely the kind of confound (here, what is actually in the
% retrieval path; in our experiments, how much context is returned and in what
% modality) that an uncontrolled leaderboard hides.

\subsection{A modern example: Hindsight}
\label{sec:hindsight}

% \begin{figure}[!t]
%   \centering
%   \includegraphics[width=\linewidth]{hindsight.png}
%   \caption{The Hindsight architecture; illustration by~\citet{hindsight}.}
%    % We hold the synthesis model (gpt-4.1-mini, $T{=}0$) and the primary
%   % judge (Gemini~3.1~Pro) fixed, and vary the retrieval budget and modality
%   % explicitly while reporting rubric audit and judge choice as
%   % evaluation-hygiene controls.}
%   \label{fig:hindsight}
% \end{figure}

As a detailed example of a strong contemporary system we use
\emph{Hindsight}~\citep{hindsight}, which partitions memory into four
epistemically distinct networks:
\begin{itemize}
\item \emph{world} network of objective facts,
\item \emph{experience} network of first-person agent experience,
\item \emph{observation} network of synthesized entity profiles, and
\item \emph{opinion} network of subjective beliefs with confidence scores and timestamps.
\end{itemize}
In this approach, ``what I \emph{know} about $X$'' can be queried separately from ``what I
\emph{think} about $X$.''

The Hindsight architecture
% , illustrated in Fig.~\ref{fig:hindsight}, 
has several components. 
The retrieval component of Hindsight, TEMPR, runs four strategies in
parallel (semantic, BM25 keyword, graph activation, and temporal-graph parsing),
fuses them with reciprocal rank fusion~\citep{cormack2009rrf}
$\text{RRF}(f) = \sum_i 1/(k + \text{rank}_i(f))$, and applies a cross-encoder
reranker. Its reflection component, CARA, processes memory in priority order
(opinions and observations before raw facts) and writes new beliefs back into
the opinion network, closing a learning loop.

Hindsight also exposes
tunable \emph{disposition parameters} (skepticism, literalism, empathy) that
shape how aggressively the agent commits to evidence during reflection. It
reports large margins on conversational memory benchmarks; e.g., on
LongMemEval~S it surpasses full-context GPT-4o by more than $23$ points using a
$20$B open model, and it tops the BEAM $10$M-token benchmark at $64.1\%$ versus
$40.6\%$ for the next system. \citet{hu2025memorysurvey} rank it first on
their aggregate Agent Memory Benchmark, and we include Hindsight as one of our
evaluated systems.

\subsection{Memory and scientific-QA benchmarks}
\label{sec:benchmarks}

The most common memory benchmarks remain
LongMemEval~\citep{wu2025longmemeval}, LoCoMo~\citep{maharana2024locomo},
BEAM~\citep{tavakoli2026beam}, and MemoryArena~\citep{he2026memoryarena}, all of
which target conversation. LoCoMo uses very long multi-session dialogues
($\sim$$300$ turns, $35$ sessions) with five question types; since long-context
models can partly solve it by ingesting the whole dialogue, scores depend
heavily on the backbone LLM. LongMemEval ($500$ questions) evaluates information
extraction, multi-session reasoning, temporal reasoning, knowledge updates, and
abstention, and is harder to solve by brute-force context. BEAM scales to
$10$M tokens and $2{,}000$ validated questions over ten abilities including
contradiction resolution. MemoryArena and its relative
Mem2ActBench~\citep{shen2026mem2actbench} measure retrieval-\emph{to-action}
grounding inside agentic tasks and expose the benchmark-to-deployment gap that
we discussed above. HaluMem~\citep{halumem} measures hallucinations {during} memory
formation, which matters because an LLM-fabricated fact written at storage time
can persist in memory storage forever.

Scientific QA benchmarks are closer to our setting but evaluate a different
object. QASA~\citep{lee2023qasa} and PeerQA~\citep{baumgartner2025peerqa} focus
on article- or document-level scientific QA, while
OpenScholar/ScholarQABench~\citep{asai2024scholarqabench} and
ResearchQA~\citep{yifei2025researchqa} evaluate multi-paper literature synthesis
and long-form scholarly answering. Our benchmark instead treats the evaluated
object as a persistent \emph{memory system}: papers are ingested into a memory
store, systems retrieve under explicit budget and modality constraints, and the
same synthesizer and judge are held fixed across memory representations. We are
not aware of a prior benchmark that uses full-paper scientific corpora across
two domains, varies retrieval budget and modality while holding the synthesizer
fixed, audits rubrics against full cited-paper text, and cross-validates
LLM-as-judge with a second frontier model and human side-by-side votes.

% =============================================================================
\section{Systems evaluated in this work}
\label{sec:systems}
% =============================================================================

We evaluate eight memory/retrieval systems plus a no-retrieval base model,
chosen to span the methodological families of Section~\ref{sec:related}.
Table~\ref{tab:systems} summarizes them; the ``Native chars'' column reports the
mean retrieved-context size per query in the uncapped run on \textsc{Paim} and
already foreshadows one of our central points: these volumes differ by more than two
orders of magnitude, so any comparison that does not control for them is
comparing budgets as much as architectures.

\begin{table}[!t]
  \centering
  \caption{Memory systems evaluated in this work. ``Native chars'' is the mean
  retrieved-context size per query in the uncapped run on \textsc{Paim};
  per-system implementation details are given below.}
  \label{tab:systems}
  \small
  \setlength{\tabcolsep}{6pt}
  \begin{tabular}{@{\extracolsep{\fill}}lp{.32\linewidth}rp{.26\linewidth}@{}}
    \toprule
    \textbf{System} & \textbf{Representation / ingest unit} & \textbf{Native chars} & \textbf{Design} \\
    \midrule
    Simple RAG       & raw chunks / 1K-char chunks       & $20{,}635$      & cosine top-$k$ baseline \\
    Theoria          & evidence + theories / full paper  & $28{,}857$      & community-routed, theory layer \\
    Mem0             & atomic facts / 6K-char chunks     & $6{,}222$       & extract-and-update \\
    Hindsight        & narrative facts / full paper item & $19{,}078$      & 4-network temporal RRF \\
    Cognee           & typed KG / full paper             & $281{,}524$     & ECL pipeline, KG completion \\
    Graphiti / Zep   & temporal graph + episodes / full paper episode & $2{,}596{,}250$ & bi-temporal, label propagation \\
    Prism            & claims + RAPTOR + links / full paper & ---          & multi-aspect typed claims \\
    Direct Read      & raw papers + filler / full corpus & $1{,}956{,}298$ & oracle-anchor diagnostic \\
    \bottomrule
  \end{tabular}
\end{table}

We group the eight systems by what they preserve and how they retrieve. We mostly used
the out-of-the-box configurations that a developer would use by default (except when
controlling for variables such as token budgets); each system is integrated through
a thin adapter in the released \texttt{scibench}
harness\footnote{Released as part of our \emph{Scientific Recall Bench}
repository, \url{https://gitlab.com/quantellence/research/scientific-recall-bench}.}
that exposes a uniform
\texttt{ingest} / \texttt{retrieve} interface and translates per-system hyperparameters
(\texttt{top\_k}, episode budgets, edge/node limits, RRF's $k$) into the
harness's per-query budget target.

\begin{enumerate}[label={\arabic*.},leftmargin=*,itemsep=3pt]
\item \textbf{Simple RAG}~\citep{lewis2020rag}, a chunk-RAG baseline. It ingests
  1K-character overlapping chunks and returns the cosine top-$k$ to an external
  synthesizer. It preserves raw text but no structure, and it is our null
  hypothesis: any memory architecture is theoretically supposed to beat it under a matched budget.

\item \textbf{Mem0}~\citep{mem0}, an extract-and-update memory. An LLM extracts
  atomic facts from $\sim$$6$K-character chunks under an ADD/UPDATE/DELETE/NOOP
  policy (the $6$K window is forced by an $8$K-token embedding cap), and an
  optional BM25 hybrid (the \texttt{fastembed} backend) is fused at retrieval
  time. Its native context is the smallest of all systems ($\sim$$6$K chars).

\item \textbf{Hindsight}~\citep{hindsight}, a narrative/temporal memory. It
  ingests each paper as an item into four parallel networks
  (world / experience / observation / opinion), retrieves with reciprocal rank
  fusion~\citep{cormack2009rrf} of semantic $+$ BM25 $+$ graph $+$ temporal
  channels, and applies a cross-encoder reranker. The server-side recall caps
  its return at $\sim$$19$K characters regardless of the requested \texttt{n},
  so it does not really honor the budget parameter; it also expects conversation
  exchanges rather than $20$K-token papers, and its $\sim$$300$KB ingest cap 
  rejected three papers.

\item \textbf{Cognee}~\citep{cognee}, a knowledge-graph memory. Its
  Extract--Cognify--Load (ECL) pipeline builds a typed Pydantic-schema KG and
  answers via \texttt{GRAPH\_COMPLETION} (internal synthesis). Its chunks are
  coarse ($\sim$$19$K characters each), so the budget grid collapses to a coarse
  $1/2/3$-chunk grid.

\item \textbf{Graphiti / Zep}~\citep{graphiti}, a bi-temporal graph memory. It
  ingests each paper as one \texttt{EpisodeType.text} episode and returns edges
  (LLM-extracted fact strings, $\sim$$200$ chars each) and nodes ($\sim$$500$
  chars each) plus, in the native run, the matched episode body, that is, a full
  $\sim$$20$K-token paper, which is why its native context reaches $2.6$M
  characters per query. The ``KG-only'' mode used in the budget sweep sets
  \texttt{include\_episodes=false} to isolate the graph from the raw-episode
  channel, and \texttt{top\_k} is applied per modality rather than globally.

\item \textbf{Theoria} and its ablation \textbf{Prism} (our systems,
  Section~\ref{sec:theoria}). Theoria ingests at the paper level, runs five LLM
  passes per chunk for multi-aspect evidence extraction, builds a RAPTOR
  collapsed tree, performs Leiden community detection, and produces theory
  statements; its BM25 sparse vectors are stored as SQLite BLOBs alongside the
  evidence and theory tables. Prism keeps the multi-aspect extraction and a
  single RAPTOR tree with typed cross-document links but answers via internal
  synthesis, so it isolates ``extraction $+$ RAPTOR'' from ``community routing
  $+$ theory aggregation.''

\item \textbf{Direct Read Anchored}, a long-context oracle. It concatenates the
  gold-source papers plus random filler up to a $500$K-token budget and runs no
  LLM at retrieval time. It is a diagnostic anchor (run on a 15-question
  subset), not a fair competitor.

\item \textbf{Base model} has no retrieval mechanism, sees no retrieved context,
and answers from parametric knowledge alone. It serves as a floor and, 
since the corpora are public arXiv papers, as a contamination probe.
\end{enumerate}
Per-system ingest cost, wall-clock, and configuration files are released in the
artifact.

% =============================================================================
\section{Datasets}
\label{sec:bench}
% =============================================================================

We release two full-text scientific QA
datasets.\footnote{Both datasets, including the paper corpora, structured
notes, questions, and audited rubrics, are available at
\url{https://huggingface.co/datasets/quantellence/srb-data}.} Both use open-access,
markdown-converted papers, an 8 question-type $\times$ 3 difficulty-tier schema,
and auditable rubrics grounded in the papers. The two corpora differ in domain
and lexical character, which lets us separate effects of
memory architectures from effects of a specific subject domain.

\textbf{\textsc{Public AI Memory} (\textsc{Paim}).} 81 full-text papers
($1.10$M words; $2.15$M tokens after MinerU markdown extraction~\citep{mineru})
on LLM agent memory, RAG, scientific agents, long-context, and adjacent
retrieval and cognitive-architecture work; in part, it is the literature surveyed
in Section~\ref{sec:related}. We also include 103 structured paper notes (under
a 10-section schema); 22 production frameworks without an arXiv mirror (e.g.,
Letta, Cognee, MemPalace, and the CrewAI/LlamaIndex/LangChain memory subsystems,
plus a pair of foundational psychology papers) are notes-only. The QA part
comprises 66 main questions across three difficulty tiers ($9$~L1, $39$~L2,
$18$~L3) plus a 10-question holdout set. Question types include factual,
mechanistic, quantitative, enumeration, conditional, cross-document, negative,
and synthesis; most questions combine two or more types.

\textbf{\textsc{Public Transformers} (\textsc{PTr}).} 252 full-text papers
(arXiv 2507.* through 2604.*) on attention mechanisms, positional encodings,
mixture-of-experts, long-context KV management, training, multimodality, and
reasoning, organized into 15 thematic clusters. \textsc{PTr} includes 98 audited
questions ($34$~L1, $41$~L2, $23$~L3), designed to cover $\sim$$48\%$ of the
corpus by source-note references and to probe the full paper-ID range. Compared
to \textsc{Paim}, its named entities---model names, kernel names, and
hyperparameters---are denser and more lexically distinctive, a property that
turns out to matter a great deal for retrieval modality
(Section~\ref{sec:ptr_results}).

\textbf{Question schema.} The three difficulty tiers capture how much of the
corpus an answer must touch. L1 questions are answerable from a single span in a
single paper (e.g., a reported score); L2 questions require a mechanism, a
multi-part enumeration, or a comparison within one or two papers; L3 questions
require synthesizing evidence across several papers (e.g., comparing thresholds
reported by three different systems). The question types are orthogonal to
the tiers: factual and quantitative questions probe precise recall, mechanistic
and conditional questions check whether qualifiers survive ingestion,
enumeration and cross-document questions target coverage, negative questions probe
calibrated abstention, and synthesis questions are intended for cross-paper aggregation.
This design deliberately stresses the parts of context restoration that compact
summaries tend to discard.

\textbf{Rubric audit.} Each question has a gold answer decomposed into
\emph{must-have facts}, each supported by a span in the corpus. We additionally
ran a 6-dimension audit by parallel reading agents, asking:
\begin{enumerate}[label=(\arabic*)]
\item is every must-have fact verifiable in the cited paper?
\item do the rubric's source pointers exist in the corpus?
\item is the answer stable (pinned to the paper, not to volatile repository
  state)?
\item can a frontier LLM answer it from parametric knowledge alone?
\item is the difficulty label still right?
\item does the rubric require precision that the cited paper does not actually claim?
\end{enumerate}
On \textsc{Paim}, the audit found that 35 questions in the first version had at
least one defect (Table~\ref{tab:audit}); these questions were rewritten and
re-audited. The same procedure was applied to \textsc{PTr}, and we release only
successfully audited questions. This audit is itself part of the contribution:
benchmark errors that are invisible at the leaderboard level (a must-have fact
that lives only outside the corpus, or a rubric that demands more precision than
the paper offers) can silently penalize systems that answer honestly.

\begin{table}[t]
  \centering
  \caption{Audit-class statistics on the first version of \textsc{Paim}
  ($76$ audited questions). Defective questions were rewritten and re-audited
  ($36/37$ clean on the first re-pass, $1$ patched again).}
  % ; the released \textsc{Paim}
  % ships the audited rubrics. The same 6-gate audit produced \textsc{PTr}'s
  % rubrics from scratch with no defects requiring a rewrite.}
  \label{tab:audit}
  \small
  \begin{tabular}{@{}clr@{}}
    \toprule
    \textbf{Class} & \textbf{Defect} & \textbf{\#} \\
    \midrule
    A & rubric fact not verifiable in the cited paper & $11$ \\
    B & dependency on internal (non-corpus) data       & $5$ \\
    C & volatile value (repository state, stars, README) & $3$ \\
    D & rubric over-specifies precision the paper does not claim & $12$ \\
    E & answerable from parametric knowledge alone (kept, flagged) & $5$ \\
    \multicolumn{2}{@{}l}{Multi-class (two classes simultaneously)} & $6$ \\
    \midrule
    \multicolumn{2}{@{}l}{\textbf{Total questions rewritten}} & $\mathbf{37}$ \\
    \bottomrule
  \end{tabular}
\end{table}

\textbf{Why full text matters.} The same systems behave very differently on full
papers than on a smaller summary corpus we previously tried. Structured-notes
ingestion sees a pre-extracted, high-density abstract; full-paper ingestion must
contend with introductions, related-work sections, numbers dispersed across
tables, redundant restatements, and inconsistent OCR around equations introduced
by the open-source MinerU PDF-to-markdown pipeline~\citep{mineru} we use for
extraction. Compact representations tend to lose qualifiers and conditions, which
are exactly what L2 and L3 questions ask about. We provide full details of our data
collection and auditing pipeline, following the W3C PROV-O
\texttt{prov:wasGeneratedBy} pattern, in the release; Section~\ref{sec:systems}
documents per-system ingestion, and Table~\ref{tab:audit} summarizes the audit
findings.

\textbf{Two examples.} To illustrate the question style, below we show two
representative questions with their gold rubrics; we follow each one through the
systems' answers in Section~\ref{sec:qualitative}.

\begin{quote}
\noindent\textbf{PQ40} (\textsc{Paim}, L2; enumeration $+$ cross-document).\\
\textbf{Q:} ``MemoryAgentBench evaluates four core competencies essential for
memory agents. List them. Which of the four is most often neglected by prior
benchmarks like LoCoMo and LongMemEval?''\\
\textbf{Gold:} the four competencies are \emph{accurate retrieval},
\emph{test-time learning}, \emph{long-range understanding}, and
\emph{selective forgetting}; selective forgetting is the most neglected---LoCoMo
and LongMemEval test recall under accumulation but do not stress-test
invalidation of stale information.\\
% \textbf{Sources:} MemoryAgentBench~\cite{hu2025memoryagentbench}, LoCoMo~\cite{maharana2024locomo}, LongMemEval~\cite{wu2025longmemeval}
\textbf{Sources:} \cite{hu2025memoryagentbench}, \cite{maharana2024locomo}, \cite{wu2025longmemeval}
\end{quote}

\begin{quote}
\noindent\textbf{TX17} (\textsc{PTr}, L2; quantitative $+$ cross-document synthesis).\\
\textbf{Q:} ``Several papers argue that sparse/linear attention only pays off
above a context-length threshold. Compare the thresholds reported by at least
three of FSA, SALS, Ring-flash-linear-2.0, HSA-UltraLong, and Kimi Linear, and
identify the most aggressive throughput claim at $1$M$+$ tokens.''\\
\textbf{Gold:} Kimi Linear is the most aggressive ($6\times$ decoding throughput
at $1$M tokens); FSA reports up to $1.25\times$ training / $1.36\times$ prefill
speedup, SALS $6.4\times$ KV compression at $4$K, Ring-flash-linear-2.0
advantages pronounced beyond $8$K, and HSA-UltraLong $>90\%$ retrieval at $16$M
tokens.\\
% FSA yan2025fsa (new, 2508.18224) · SALS mu2025sals (new, 2510.24273) · Ring-flash-linear-2.0 lingteam2025ringflashlinear (new, 2510.19338) · HSA-UltraLong hu2025ultralong (new, 2511.23319) · Kimi Linear kimiteam2025kimilinear (new, 2510.26692)
\textbf{Sources:} \cite{yan2025fsa}, \cite{mu2025sals}, \cite{lingteam2025ringflashlinear}, \cite{hu2025ultralong}, \cite{kimiteam2025kimilinear}
\end{quote}

% =============================================================================
\section{Theoria: a three-layer scientific-memory system}
\label{sec:theoria}
% =============================================================================

\begin{figure}[!t]
  \centering
  \includegraphics[width=\linewidth]{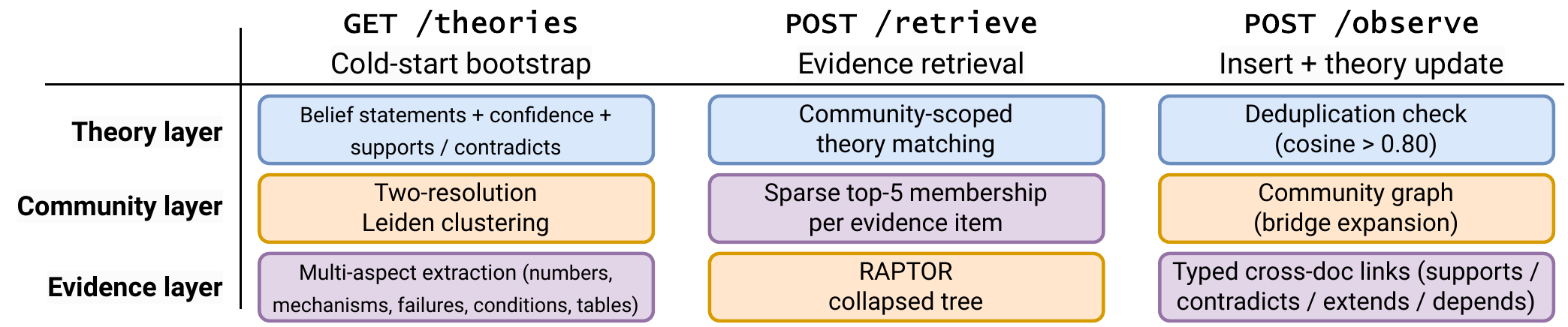}
  \caption{The architecture of \emph{Theoria}. The three-layer
  evidence~$+$~community~$+$~theory stack shares the same store with three
  agent endpoints: \texttt{GET /theories} for cold-start bootstrap,
  \texttt{POST /retrieve} for community-routed evidence retrieval, and
  \texttt{POST /observe} for inserting new findings (which can update or
  contradict existing theories).}
   % The counts shown are after ingesting the 81-paper \textsc{Paim} corpus.}
  \label{fig:theoria}
\end{figure}

We introduce \emph{Theoria}, our structured scientific-memory system and one of
the systems evaluated below. Theoria is built around the hypothesis that scientific
memory needs {both} fine-grained evidence retrieval (table values, precise
conditions, failure modes) {and} coarse-grained structure for cross-paper
synthesis (which papers belong together, where their claims disagree). Most
production systems pick one: chunk RAG~\citep{lewis2020rag} preserves text but
not structure, while knowledge-graph completion in Cognee~\citep{cognee} imposes
structure but loses raw-text fidelity. 

We therefore view Theoria both as a
systems contribution and as a test subject for the evaluation protocol: if structure
helps, it should show up under matched retrieval budgets. Theoria stacks three
layers so that evidence retrieval and theory-level cross-paper aggregation share
the same store but expose distinct interfaces (Fig.~\ref{fig:theoria}).

\textbf{Evidence layer.} Theoria performs multi-aspect extraction over
section-aware $\sim$$6$K-char chunks, with five LLM passes that extract
\textsc{numbers}, \textsc{mechanisms}, \textsc{failures}, \textsc{conditions},
and \textsc{tables} as self-contained claims. A RAPTOR collapsed tree~\citep{sarthi2024raptor}
builds one summary level above the leaves via Ward agglomerative
clustering. Typed cross-document links (\textsc{supports} / \textsc{contradicts}
/ \textsc{extends} / \textsc{depends-on}) are discovered by filtering: each new
claim's top-5 embedding-nearest existing claims are passed to an LLM
disambiguator, and typed relations become rows of a \texttt{links} table. After
ingesting \textsc{Paim}, Theoria holds $26{,}782$ evidence items plus RAPTOR
summary nodes and $1{,}275$ typed cross-document links.

\textbf{Community layer.} We run two-resolution Leiden
clustering~\citep{traag2019leiden} on the evidence graph (with edges between
embedding-near items) and assign items to fine communities, weighted by cosine
similarity to community centroids. Community-to-community edges drive expansion
at retrieval time. On \textsc{Paim}, this yields $\sim$$70$ coarse and
$\sim$$470$ fine communities, with average degree $\sim$$12$. The intuition is
that a research question rarely lands on a single isolated fact; routing through
communities lets retrieval pull in the cluster of related evidence that a
synthesis answer needs.

\textbf{Theory layer.} ``Theories'' are natural-language statements anchored to
a community, carried at \textsc{low} / \textsc{medium} / \textsc{high}
confidence, promoted by supporting-evidence count and demoted by contradiction
count. When a new finding arrives, only theories anchored to its top-5
communities (plus their direct graph neighbors) are considered as candidates,
narrowing the comparison from all theories to typically $10$--$40$. Cosine
similarity above $0.25$ counts as a match; on no match (and below the
deduplication threshold of $0.80$) the finding seeds a new theory. On
\textsc{Paim} this produces $\sim$$637$ theories.

These layers support three agent endpoints (Fig.~\ref{fig:theoria}):
\begin{enumerate}[label=(\roman*)]
\item \texttt{GET /theories} returns the theory table filtered by confidence and
  grouped by community---an agent's cold-start bootstrap (``what does this corpus
  believe?'');
\item \texttt{POST /retrieve} embeds the query, picks the top-5 fine communities,
  optionally expands to neighbors, scores by entropy-weighted cosine across
  RAPTOR levels, and returns the top-$k$ deduplicated items plus their one-hop
  linked items;
\item \texttt{POST /observe} ingests a new finding: extract evidence, assign
  communities, and update existing theories or seed new beliefs.
\end{enumerate}
Below, we evaluate only Theoria's \texttt{POST /retrieve} endpoint under
matched-budget single-shot QA. This is a deliberately conservative evaluation:
it tests whether community-routed evidence and theory-linked retrieval improve
answer quality, but it does not use the full agentic loop. The cold-start
\texttt{GET /theories} endpoint and the iterative \texttt{POST /observe} loop---
the parts of Theoria most unlike RAG, and the ones that have proven most useful
in our internal use---are \emph{not} exercised by one-shot QA; evaluating them
properly requires a sequential research-agent benchmark, which we leave to
future work.

\textbf{Prism (ablation).} Our second system, \emph{Prism}, shares Theoria's
multi-aspect extraction but replaces the community and theory layers with a
single RAPTOR tree and typed links, returning answers via internal synthesis
rather than external retrieval. Prism therefore isolates ``multi-aspect
extraction $+$ RAPTOR'' from ``community routing $+$ theory aggregation.''
Because internal synthesis has no measurable retrieved-context volume, Prism
does not participate in the budget sweeps.

\section{Experimental evaluation}
\label{sec:eval}

% =============================================================================
\subsection{Experimental setup}
\label{sec:setup}
% =============================================================================

We run all memory systems on \textsc{Paim} (81 papers, 66 questions) and on
\textsc{PTr} (252 papers, 98 questions). External-synthesis systems use
gpt-4.1-mini at temperature $0$ as the shared synthesizer; internal-synthesis
systems (Cognee, Prism) return their own answer using whatever model they invoke
internally. The primary judge is Gemini~3.1~Pro, scoring each answer against the
gold rubric on five dimensions: accuracy, completeness, specificity,
hallucination avoidance, and retrieval quality. We report the \emph{no-retrieval
composite}, the mean of the first four dimensions on a $0$--$10$ scale, so that
internal- and external-synthesis systems are scored on the same footing, and we
use the audited rubrics of Section~\ref{sec:bench} throughout.

\textbf{Native and budget-targeted configurations.} There are two
configurations. In the \emph{native} configuration retrieval is uncapped and
each system returns whatever it would out of the box. In the
\emph{budget-targeted} configuration we re-query the already-ingested memory at
three character targets $B \in \{10\text{K}, 30\text{K}, 50\text{K}\}$ by
changing only retrieval settings (\texttt{top\_k} or its equivalent); the
corpus, the stored representation, and the synthesis pipeline are unchanged.
This separation is deliberate: the native track measures out-of-the-box product
behavior, while the budget-targeted track measures evidence-selection efficiency
at a fixed context cost. We calibrate the retrieval settings from each system's
mean characters per retrieved unit on the native run (Theoria $1{,}151$,
Hindsight $432$, Mem0 $292$, Cognee $18{,}768$, Graphiti edges$+$nodes $223$,
Simple RAG $999$).

\textbf{Two ablation variants.} For Graphiti, the budgeted runs additionally disable
episode retrieval (\texttt{include\_episodes\allowbreak=false}); the system then returns
only edges (LLM-extracted fact strings, $\sim$$200$ chars each) and nodes
($\sim$$500$ chars each), because each episode is a $\sim$$20$K-token full-text
paper that would by itself blow any budget. This ``KG-only'' Graphiti is an
ablation that isolates the graph representation from the raw-episode channel.
Rows marked ``$+$ BM25'' are the sparse--dense hybrids: BM25 fused with the
existing dense retrievers by reciprocal rank fusion at $k=60$, computed without
re-ingesting the corpus.

% \textbf{Systems excluded from the budget sweep.} 
Three systems cannot take part
in the budget sweep: Prism (internal synthesis, no measurable retrieved-context
volume), Direct Read (an oracle anchor at $\sim$$2$M chars/query that exceeds
every budget, run on its 15-question subset), and the base model (no retrieval).

% =============================================================================
\subsection{Results on \textsc{Paim}: native and budget-targeted retrieval}
\label{sec:paim_results}
% =============================================================================

\begin{table}[!t]\centering
  \caption{Experimental results on \textsc{Paim} (Gemini 3.1 Pro judge, 0--10 scale).
   % \emph{Native (66q)}: full-text run with no retrieval cap; per-tier $n$ are L1$=9$, L2$=39$, L3$=18$. \emph{$B\!\approx\!10/30/50$K}: budget-targeted retrieval sweep (\texttt{top\_k} chosen so the retrieved context approx. fits the target); ``chars'' is the achieved mean retrieved characters per query, ``score'' is the corresponding no-retrieval composite.
   }
   % The three rows marked ``$+$ BM25'' are the BM25 hybrid (BM25 over RRF $k\!=\!60$, fused with the existing dense retrievers).}
  \label{tab:paim}
  \scriptsize
  \setlength{\tabcolsep}{3pt}
  \begin{tabular*}{\linewidth}{@{\extracolsep{\fill}}lccccccccccc@{}}
    \toprule
    & \multicolumn{5}{c}{\textbf{Native (66 questions)}}
    & \multicolumn{2}{c}{\textbf{$B\!\approx\!10$}K}
    & \multicolumn{2}{c}{\textbf{$B\!\approx\!30$}K}
    & \multicolumn{2}{c}{\textbf{$B\!\approx\!50$}K} \\
    \cmidrule(lr){2-6}\cmidrule(lr){7-8}\cmidrule(lr){9-10}\cmidrule(lr){11-12}
    \textbf{System}
    & \textbf{Avg} & \textbf{L1} (9q) & \textbf{L2} (39q) & \textbf{L3} (18q) & \textbf{chars}
    & \textbf{score} & \textbf{chars}
    & \textbf{score} & \textbf{chars}
    & \textbf{score} & \textbf{chars} \\
    \midrule
    Theoria                       & $6.82$          & $8.14$          & $6.64$          & $6.54$          & $29$K          & $6.32$           & $16$K       & $7.19$           & $41$K       & $7.60$           & $60$K       \\
    \quad $+$ BM25 hybrid         & $6.97$          & $8.09$          & $6.51$          & $\mathbf{7.22}$ & $28$K          & $\mathbf{6.70}$  & $16$K         & $\mathbf{7.41}$  & $40$K         & $7.34$           & $57$K         \\
    Mem0                          & $5.65$          & $8.44$          & $5.35$          & $4.90$          & $6$K           & $6.43$           & $11$K       & $6.97$           & $32$K       & $7.40$           & $53$K       \\
    \quad $+$ BM25 hybrid         & $5.98$          & $8.33$          & $5.54$          & $5.74$          & $6$K           & $6.54$           & $11$K         & $7.13$           & $33$K         & $7.48$           & $54$K         \\
    Simple RAG                    & $\mathbf{7.22}$ & $8.28$          & $\mathbf{7.14}$ & $6.86$          & $21$K          & $6.27$           & $11$K       & $7.25$           & $31$K       & $7.58$           & $52$K       \\
    \quad $+$ BM25 hybrid         & $7.15$          & $\mathbf{8.47}$ & $6.77$          & $7.03$          & $21$K          & $6.64$           & $10$K         & $7.30$           & $32$K         & $\mathbf{7.77}$  & $53$K         \\
    Hindsight                     & $6.81$ & $7.86$ & $6.44$ & $7.10$ & $19$K                & $6.63$           & $19$K       & $6.53$           & $19$K       & $6.72$           & $19$K       \\
    Cognee                        & $6.20$ & $7.14$ & $6.13$ & $5.89$ & $282$K               & $5.16$           & $19$K       & $5.28$           & $37$K       & $5.45$           & $56$K       \\
    Graphiti                   & $8.04$ & $8.49$ & $7.84$ & $8.24$ & $2{,}552$K              & $5.05$           & $21$K       & $5.27$           & $59$K       & $5.19$           & $96$K       \\
    \midrule
    % Graphiti (native)$^{\ddagger}$ & $8.04$ & $8.49$ & $7.84$ & $8.24$ & $2{,}552$K          & \multicolumn{6}{l}{Native retrieval; not in budget sweep} \\
    Direct Read (on 15q)      & $6.67$ & $6.30$ & $6.15$ & $7.55$ & $1{,}968$K           & \multicolumn{6}{l}{Oracle anchor; budgets exceeded} \\
    Prism (ablation)              & $6.20$ & $8.14$ & $5.92$ & $5.83$ & ---                  & \multicolumn{6}{l}{Internal synthesis (no retrieval volume)} \\
    Base model                    & $2.64$ & $1.50$ & $2.51$ & $3.49$ & $0$                  & $2.64$           & $0$         & $2.64$           & $0$         & $2.64$           & $0$         \\
    \bottomrule
  \end{tabular*}
  % \smallskip
  % \scriptsize{$^{\ddagger}$Graphiti (native) uses \texttt{include\_episodes=true}, so each chunk adds a $\sim$$20$K-token paper; Graphiti~(KG) uses \texttt{include\_episodes=false} (edges+nodes only) and is the budget-comparable variant.}
\end{table}

\begin{figure}[!t]
  \centering
  \includegraphics[width=\linewidth]{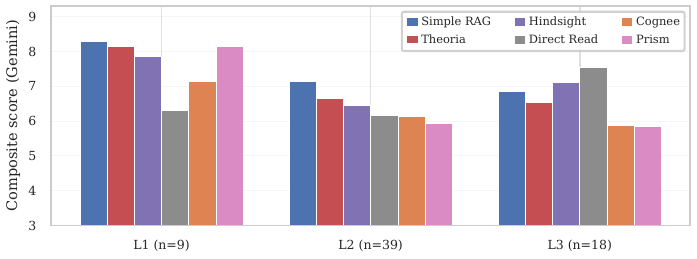}
  \caption{Native \textsc{Paim} scores by difficulty tier (Gemini judge,
  audited rubrics). L1 is largely saturated; L2 fans the systems out; on L3 the
  long-context Direct Read anchor ($\sim$$2$M chars/query) leads, followed by
  Hindsight and Simple RAG. These are the ``Native'' columns of
  Table~\ref{tab:paim}.}
  \label{fig:native_tier}
\end{figure}

Tables~\ref{tab:paim} and~\ref{tab:ptr} report the full results on \textsc{Paim}
and \textsc{PTr}; Figure~\ref{fig:native_tier} shows the \textsc{Paim} results
graphically, and Figure~\ref{fig:budget} plots the budget sweeps.
We find three patterns in Table~\ref{tab:paim}.

First, \emph{L1 is largely saturated.} Every retrieval-augmented system other than
Direct Read reaches $\geq 7.7$ on L1 in the native run, with little spread;
systems separate only from L2 downward. Basic factual recall is therefore no
longer a useful discriminator among memory architectures---it mainly verifies
that ingestion and retrieval are wired up at all.

Second, \emph{native L3 is dominated by raw context volume.} The two highest native
L3 scores belong to the two highest volume systems. Graphiti (native, $8.24$ on
L3) returns $\sim$$2.55$M chars/query because it ingests each paper as one
episode, so every matched episode adds the whole $\sim$$20$K-token paper into
synthesis; Direct Read ($7.55$) is a diagnostic with $\sim$$2$M chars/query of
gold sources plus random filler. Without budget control, neither gain can be
attributed to architecture: a system that hands the synthesizer two million
characters of relevant raw text is, in effect, a semantically filtered
long-context reader, not evidence that temporal-graph memory is intrinsically
better.

Third, \emph{under budget control, the architectural premium over chunk RAG is
small or negative on \textsc{Paim}.} At the fair-comparison anchor $B\approx
30$K (the natural volume of several systems), Simple~RAG $7.25 \approx$ Theoria
$7.19 >$ Mem0 $6.97 >$ Hindsight $6.53 >$ Cognee $5.28 \approx$ Graphiti~(KG)
$5.27$. The KG-only Graphiti row lands at the bottom: its native wins were
entirely due to the raw episode bodies, not the graph. Simple RAG, Theoria, and
Mem0 form a tight upper cluster that scales monotonically with budget
($\sim$$6.3$ at $10$K, $\sim$$7.2$ at $30$K, $\sim$$7.6$ at $50$K), staying
within $0.3$ points of one another. Hindsight saturates at $\sim$$19$K chars
regardless of \texttt{n} because its server caps recall, so its three budget
columns are nearly identical; Cognee's chunks are $\sim$$19$K each, mapping the
budget grid onto $\texttt{top\_k}=1/2/3$, too coarse for a real curve.

Figure~\ref{fig:native_tier} visualizes per-tier scores for external synthesis systems:
L1 is saturated, L2 distinguishes between the systems, and on L3 the long-context
Direct Read anchor leads, followed by Hindsight and Simple
RAG, which is again a volume effect rather than an architecture effect.

\textbf{BM25 hybrid on \textsc{Paim}.} For Mem0, Simple RAG, and Theoria we
additionally evaluate a BM25-over-RRF hybrid~\citep{cormack2009rrf,robertson2009bm25}
that fuses the sparse and dense retrievers ($k=60$) without re-ingesting the
corpus; these are the rows marked ``$+$ BM25 hybrid'' in Table~\ref{tab:paim}.
On \textsc{Paim} the lift is uneven. In the native run it is small ($+0.15$ for
Theoria, $+0.33$ for Mem0, $-0.07$ for Simple RAG); under budget control BM25 is
most useful at low and mid budgets ($+0.4$ at $10$K, $+0.1$ to $+0.2$ at $30$K)
and turns slightly negative at $50$K for Theoria ($-0.26$). The picture on
\textsc{Paim} is that retrieval modality is not decisive and architecture
matters less than context budget. As we show next, \textsc{PTr} tells a very
different story about modality.

\textbf{Granularity caveats.} Three integrations diverge from their intended
granularity (Section~\ref{sec:systems}): Hindsight is fed whole papers rather
than conversation exchanges, and its server cap rejects the largest three, Mem0
chunks at an arbitrary $6$K window, and Graphiti's episode is meant for short
snippets rather than whole papers. We treat these as evaluation facts rather than
defects, since they are the out-of-the-box choices a developer actually
faces: a system has a specific granularity, size and semantic coherence of
a unit it expects at ingestion, and violating their default parameters
can change both quality and cost. Qualitative examples follow in
Section~\ref{sec:qualitative}; we present more in Appendix~\ref{app:examples}.

% =============================================================================
\subsection{Results on \textsc{PTr}: a different domain and the BM25 hybrid effect}
\label{sec:ptr_results}
% =============================================================================

% PTr master results table (replaces the PTr columns of table 4 in main_v3.tex).
% Combines: Phase 2 overall + per-tier scores on the full 98-question set
% (48 baseline questions + 50 expansion questions, 252 papers, audited rubrics)
% and the budget-targeted retrieval sweep at 10/30/50K characters.
% Per-tier is reported at the headline B=50K budget.
% Scores: Gemini 3.1 Pro judge, no-retrieval composite (mean of accuracy /
% completeness / specificity / hallucination-avoidance), 0--10 scale.
% Mean retrieved characters are computed as the per-question weighted mean
% over the 48-question baseline run (phase2) and the 50-question expansion
% run (phase2_v22), the same protocol as for \textsc{Paim} in Table~\ref{tab:paim}.
\begin{table}[!t]
  \centering
  \caption{Experimental results on \textsc{PTr} (Gemini 3.1 Pro judge, 0--10 scale).
  % ; 252 papers, 98 questions: 48 baseline + 50 coverage-targeted expansion. \emph{Headline (98q at $B\!=\!50$K)}: overall mean, per-tier scores ($n_{\text{L1}}\!=\!34$, $n_{\text{L2}}\!=\!41$, $n_{\text{L3}}\!=\!23$), and achieved mean retrieved characters per query.
  % % No separate uncapped ``native'' run was performed for \textsc{PTr}, so the headline ``chars'' column repeats the $B\!\approx\!50$K column on the right. 
  % \emph{$B\!\approx\!10/30/50$K}: budget-targeted retrieval sweep over the 98 questions; ``score'' is the no-retrieval composite at the target, ``chars'' is the achieved mean retrieved characters per query. Rows marked ``$+$ BM25'' use BM25 hybrid retrieval (sparse $+$ dense, RRF fusion at $k\!=\!60$); dense rows use the dense-only baseline. Bold marks the per-column best overall and per tier; bold in the budget block marks the per-column best score at each budget.
  }
  \label{tab:ptr}
  \scriptsize
  \setlength{\tabcolsep}{3pt}
  \begin{tabular*}{\linewidth}{@{\extracolsep{\fill}}lccccccccccc@{}}
    \toprule
    & \multicolumn{5}{c}{\textbf{Native (98 questions)}}
    & \multicolumn{2}{c}{\textbf{$B\!\approx\!10$}K}
    & \multicolumn{2}{c}{\textbf{$B\!\approx\!30$}K}
    & \multicolumn{2}{c}{\textbf{$B\!\approx\!50$}K} \\
    \cmidrule(lr){2-6}\cmidrule(lr){7-8}\cmidrule(lr){9-10}\cmidrule(lr){11-12}
    \textbf{System}
    & \textbf{Avg} & \textbf{L1} (34q) & \textbf{L2} (41q) & \textbf{L3} (23q) & \textbf{chars}
    & \textbf{score} & \textbf{chars}
    & \textbf{score} & \textbf{chars}
    & \textbf{score} & \textbf{chars} \\
    \midrule
    Theoria $+$ BM25     & $8.67$          & $9.75$          & $\mathbf{8.77}$ & $7.09$          & $26$K & $8.21$          & $15$K       & $\mathbf{8.96}$ & $36$K       & $9.21$          & $50$K       \\
    Mem0 $+$ BM25        & $8.10$          & $9.54$          & $7.83$          & $6.22$          & $7$K  & $\mathbf{8.42}$ & $11$K       & $8.88$          & $34$K       & $\mathbf{9.24}$ & $57$K       \\
    Simple RAG $+$ BM25  & $\mathbf{8.74}$ & $\mathbf{9.79}$ & $8.70$          & $\mathbf{7.37}$ & $21$K & $7.94$          & $11$K       & $8.88$          & $32$K       & $9.22$          & $53$K       \\
    \midrule
    Theoria (dense)      & $8.23$          & $9.70$          & $8.23$          & $6.22$          & $28$K & $7.92$          & $16$K       & $8.54$          & $39$K       & $8.96$          & $55$K       \\
    Mem0 (dense)         & $7.73$          & $9.60$          & $7.16$          & $5.72$          & $7$K  & $8.21$          & $11$K       & $8.74$          & $33$K       & $8.95$          & $55$K       \\
    Simple RAG (dense)   & $8.31$          & $9.66$          & $8.07$          & $6.64$          & $21$K & $7.83$          & $10$K       & $8.45$          & $31$K       & $8.72$          & $52$K       \\
    \midrule
    Base model           & $1.78$          & $1.76$          & $1.66$          & $2.06$          & ---   & $1.89$          & $0$         & $1.81$          & $0$         & $1.70$          & $0$         \\
    \bottomrule
  \end{tabular*}
  \smallskip
  % \scriptsize{The 48 baseline questions cover $28\%$ of the 252-paper corpus by source-notes references; the 50 expansion questions push coverage to $48\%$ over the full 98-question set. Hindsight, Cognee, and Graphiti participate on \textsc{Paim} only. The $+$BM25 row pattern is a ``free re-index'' over the same dense store ($\sim$$80$~s of local CPU, no LLM calls); the difference vs.\ the dense row above isolates the lexical-retrieval contribution. The three hybrid systems tie within $0.03$ points at $B\!=\!50$K despite ingest costs differing by orders of magnitude (Theoria: \$30 of LLM extraction $+\sim$$10$h wall-clock; Mem0: hours of fact extraction; Simple RAG: $\sim$$1.5$~min of embedding).}
\end{table}

\textsc{PTr} has 252 full-text papers and 98 audited questions on the modern
Transformer / long-context / multimodal literature; as noted, its named entities
are denser and more lexically distinctive than \textsc{Paim}'s. Due to
budget constraints we first compared all systems on a subset of \textsc{PTr}
size-matched to \textsc{Paim} (Table~\ref{tab:ptr_subset}), then evaluated the
three strongest systems---Theoria, Mem0, and Simple RAG---and their three BM25
hybrid variants on the full corpus. Table~\ref{tab:ptr} reports the full-corpus
results, and the middle panel of Fig.~\ref{fig:budget} shows the budget curves.

% PTr 80-paper / 48-question size-matched subset of PTr (formerly "Phase 1").
% Same evaluation protocol as Table~\ref{tab:ptr}, but on a subset of PTr that matches
% the size of \textsc{Paim} ($\sim$2.17M tokens, 80 papers) so the cross-corpus
% comparison is at matched scale. The subset additionally includes Hindsight,
% Cognee, and Graphiti (which did not run on full PTr).
% Source: scibench_runs7/public_transformers_subset81_budget_{10k,30k,50k}.
% Scores: Gemini 3.1 Pro judge, no-retrieval composite, 0--10 scale.
\begin{table}[!t]\centering
  \caption{Experimental results on the 80-paper subset of \textsc{PTr} ($48$q, Gemini 3.1 Pro judge).
  % \emph{Headline (48q at $B\!=\!50$K)}: overall mean, per-tier scores ($n_{\text{L1}}\!=\!16$, $n_{\text{L2}}\!=\!20$, $n_{\text{L3}}\!=\!12$), and achieved mean retrieved characters per query (which equal the $B\!\approx\!50$K block on the right because no separate uncapped run was performed). \emph{$B\!\approx\!10/30/50$K}: budget-targeted retrieval sweep over the 48 questions; ``score'' is the no-retrieval composite at the target, ``chars'' is the achieved mean retrieved characters per query. Only Mem0 has a BM25 hybrid variant on this subset; Hindsight, Cognee and Graphiti are included here but do not run on full \textsc{PTr}. Bold marks the per-column best overall and per tier; bold in the budget block marks the per-column best score at each budget.
  }
  \label{tab:ptr_subset}
  \scriptsize
  \setlength{\tabcolsep}{4pt}
  \begin{tabular*}{\linewidth}{@{\extracolsep{\fill}}lccccccccccc@{}}
    \toprule
    & \multicolumn{5}{c}{\textbf{Headline (48 questions)}, $B\!=\!50$K}
    & \multicolumn{2}{c}{\textbf{$B\!\approx\!10$}K}
    & \multicolumn{2}{c}{\textbf{$B\!\approx\!30$}K}
    & \multicolumn{2}{c}{\textbf{$B\!\approx\!50$}K} \\
    \cmidrule(lr){2-6}\cmidrule(lr){7-8}\cmidrule(lr){9-10}\cmidrule(lr){11-12}
    \textbf{System}
    & \textbf{Overall} & \textbf{L1} (16q) & \textbf{L2} (20q) & \textbf{L3} (12q) & \textbf{chars}
    & \textbf{score} & \textbf{chars}
    & \textbf{score} & \textbf{chars}
    & \textbf{score} & \textbf{chars} \\
    \midrule
    Mem0 $+$ BM25        & $8.60$          & $9.67$          & $8.05$          & $8.08$          & $55$K & $\mathbf{8.02}$ & $11$K       & $\mathbf{8.56}$ & $33$K       & $8.60$          & $55$K       \\
    \midrule
    Theoria              & $8.45$          & $9.88$          & $\mathbf{8.24}$ & $6.92$          & $54$K & $7.27$          & $16$K       & $8.36$          & $38$K       & $8.45$          & $54$K       \\
    Mem0                 & $\mathbf{8.72}$ & $\mathbf{9.97}$ & $8.04$          & $\mathbf{8.19}$ & $53$K & $6.56$          & $10$K       & $8.33$          & $32$K       & $\mathbf{8.72}$ & $53$K       \\
    Simple RAG           & $8.17$          & $9.64$          & $7.86$          & $6.73$          & $52$K & $7.33$          & $10$K       & $8.02$          & $31$K       & $8.17$          & $52$K       \\
    \midrule
    Hindsight            & $7.07$          & $9.05$          & $6.39$          & $5.56$          & $18$K & $7.23$          & $18$K       & $7.22$          & $18$K       & $7.07$          & $18$K       \\
    Cognee               & $5.97$          & $8.25$          & $5.62$          & $3.50$          & $51$K & $5.27$          & $16$K       & $5.70$          & $33$K       & $5.97$          & $51$K       \\
    Graphiti (KG)        & $7.99$          & $8.50$          & $7.67$          & $7.83$          & $96$K & $7.61$          & $21$K       & $7.87$          & $58$K       & $7.99$          & $96$K       \\
    \midrule
    Base model           & $1.99$          & $1.78$          & $1.82$          & $2.56$          & $0$   & $1.99$          & $0$         & $1.99$          & $0$         & $1.99$          & $0$         \\
    \bottomrule
  \end{tabular*}
  % \smallskip
  % \scriptsize{The 48 questions are TX1--TX48 (the ``baseline'' subset of \textsc{PTr}, before the 50-question coverage-targeted expansion that brings the full \textsc{PTr} to $98$ questions). Subset chosen greedily to match \textsc{Paim} at $\sim$$2.17$M tokens while covering the source-notes of all 48 baseline questions. The base-model row repeats the $30$K-budget value across budgets because the no-retrieval baseline does not depend on the retrieval budget.}
\end{table}

\begin{figure}[!t]
  \centering\setlength{\tabcolsep}{2pt}
  \begin{tabular*}{\linewidth}{P{.38\linewidth}P{.3\linewidth}P{.3\linewidth}}
    \multicolumn{3}{c}{\includegraphics[width=\linewidth]{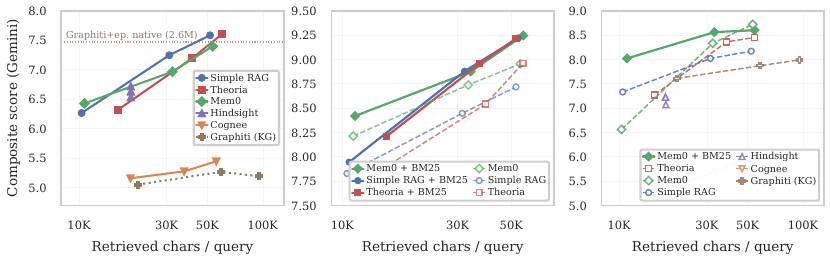}} \\[-2pt]
    \footnotesize \emph{(a)} \textsc{Paim} (66 questions) &
    \footnotesize \emph{(b)} \textsc{PTr} (98 questions) &
    \footnotesize \emph{(c)} \textsc{PTr} 80-paper subset (48q) \\
  \end{tabular*}
  \caption{Budget sweeps: composite score vs.\ mean retrieved characters per
  query (log $x$-axis). \emph{(a)~\textsc{Paim}}: Simple RAG, Theoria, and Mem0
  form a tight upper cluster that scales monotonically with budget; Hindsight is
  a single point repeated because its server ignores the budget knob; Cognee and
  Graphiti's KG output are flat and low. The dotted line marks Graphiti's native
  score with episodes ($2.6$M chars). \emph{(b)~\textsc{PTr}} (252 papers, 98
  questions): filled markers / solid lines are hybrid retrieval (BM25$+$dense,
  RRF $k=60$); open markers / dashed lines are dense-only. 
  % The three hybrids tie   within $0.03$ at $50$K. 
  \emph{(c)~\textsc{PTr} 80-paper subset} (size-matched
  to \textsc{Paim}, 48 questions; Table~\ref{tab:ptr_subset}): the same six dense
  systems as in (a), plus the Mem0$+$BM25 variant.}
  % ; the qualitative picture from
  % \textsc{Paim} replicates, and Mem0$+$BM25 lifts the leaderboard markedly at
  % $10$K and $30$K.}
  \label{fig:budget}
\end{figure}

Again we make three observations.

First, \emph{dense retrieval is tighter, but architectural advantages persist.} At
$B\approx 50$K, Theoria $8.96 \approx$ Mem0 $8.95 >$ Simple RAG $8.72$:
structured and extracted memories pull ahead of chunk RAG, though the spread is
small. Unlike \textsc{Paim}, the gap does not vanish under dense retrieval.

Second, \emph{BM25 lifts every evaluated system, and the three hybrids converge.}
Adding BM25 over dense retrieval gives $+0.50$ for Simple RAG at $50$K, $+0.29$
for Mem0, and $+0.25$ for Theoria; the lift is largest where the dense-only L3
score was lowest, suggesting that BM25 recovers exactly what dense cosine
retrieval misses. At $50$K the three hybrids are essentially tied
(Mem0$+$BM25 $9.24$, Simple~RAG$+$BM25 $9.22$, Theoria$+$BM25 $9.21$), even
though their ingest costs differ by orders of magnitude (Theoria: $\sim$\$30 of
LLM extraction $+$ $\sim$$10$h wall-clock; Mem0: hours of fact extraction;
Simple RAG: $\sim$$1.5$ minutes of embedding). The convergence is on
\emph{average} overall score only; tier-level behavior, latency, cost,
provenance, and agentic interfaces all still differ.

Third, \emph{we find tier-level specialization.} At $B\approx 50$K, Theoria$+$BM25 wins
L1 ($9.96$) and ties Simple~RAG$+$BM25 on L2 ($9.21$), but Mem0$+$BM25 wins L3
($8.63$), so a question-conditioned ensemble of memory systems could in
principle raise the ceiling further; we quantify both this headroom and the
difficulty of realizing it in Section~\ref{sec:discussion}. The dense rows show
the same but weaker pattern, with Theoria leading on L1/L2 and Mem0 on L3.

Why does BM25 help so much more here than on \textsc{Paim}? We believe the key is \textsc{PTr}'s
lexical structure: named methods, kernel names, and model identifiers, often
phrased exactly as in the paper. Dense embeddings collapse synonyms
but also conflate near-neighbors (e.g., different attention variants), whereas
BM25 preserves the surface form and can pull the exact rare phrase a question
hinges on. Table~\ref{tab:ptr_subset} shows the size-matched 80-paper / 48-question subset,
which we ran before the full corpus to compare domains at the same scale as
\textsc{Paim} and to choose systems for the full run. The subset was selected
greedily so that its token count matches \textsc{Paim} ($\sim$$2.17$M tokens) and
the source notes of the 48 baseline questions are covered. At the $30$K
fair-comparison anchor we observed Theoria $8.36$, Mem0 $8.33$, Simple RAG
$8.02$, Graphiti~(KG) $7.87$, Hindsight $7.22$, and Cognee $5.70$: the
qualitative picture from \textsc{Paim} replicates, except that here the
architectural systems (Theoria, Mem0) edge ahead of Simple RAG by $0.3$--$0.5$
points instead of tying, which is why those three were chosen for the full run.

The subset also tells a more nuanced story about scaling the corpus: comparing
the same 48 baseline questions on the 80-paper subset vs.\ the full 252-paper
corpus (same questions, more papers), dense-only Theoria moves $8.45\to8.71$
($+0.26$), Mem0 $8.72\to8.52$ ($-0.20$), and Simple RAG $8.17\to8.08$ ($-0.09$).
This directional pattern is consistent with candidate pool dilution that 
pre-aggregation absorbs but atomic fact extraction does not; the combined
98-question numbers in Table~\ref{tab:ptr} therefore mix this scale effect
with the difficulty of the 50 expansion questions, so we report both views
and avoid attributing the leaderboard change to corpus growth alone. 
Qualitative examples follow in Section~\ref{sec:qualitative}; more are in Appendix~\ref{app:examples}.

% =============================================================================
\subsection{Statistical significance of our main results}
\label{sec:significance}
% =============================================================================

Since the question counts are modest, we quantify the uncertainty of the
main differences with a {paired bootstrap} over questions:
we resample question ids with replacement ($B=10{,}000$), recompute the mean
per-question score difference each time, and report the percentile $95\%$
confidence interval. Pairing is over the common question set since every system
answers the same questions; scores are the Gemini composite used throughout.
This analysis re-aggregates the same judge scores reported elsewhere and
involves no new inference. Table~\ref{tab:bootstrap} shows the results.

\begin{table}[!t]
  \centering
  \caption{Paired bootstrap $95\%$ confidence intervals for the main contrasts
  (Gemini composite, $B=10{,}000$ resamples over questions).
  $\Delta$ is the mean per-question score difference, $n$ is the number of
  common questions, $^{*}$ marks an interval that excludes $0$.}
  \label{tab:bootstrap}
  \footnotesize
  \begin{tabular}{@{}lrlr@{}}
    \toprule
    \textbf{Contrast} & \textbf{$\Delta$} & \textbf{$95\%$ CI} & \textbf{$n$} \\
    \midrule
    \multicolumn{4}{@{}l}{\emph{\textsc{Paim}, budget-targeted @ $30$K (the fair-comparison anchor)}}\\
    Simple RAG $-$ Theoria          & $+0.01$ & $[-0.52,\,+0.53]$        & $65$ \\
    Theoria $-$ Mem0                & $+0.22$ & $[-0.28,\,+0.74]$        & $65$ \\
    Simple RAG $-$ Mem0             & $+0.28$ & $[-0.18,\,+0.76]$        & $66$ \\
    Simple RAG $-$ Hindsight        & $+0.72$ & $[+0.18,\,+1.30]^{*}$    & $66$ \\
    Simple RAG $-$ Cognee           & $+1.97$ & $[+1.29,\,+2.65]^{*}$    & $66$ \\
    Simple RAG $-$ Graphiti (KG)    & $+1.98$ & $[+1.16,\,+2.84]^{*}$    & $66$ \\
    \addlinespace
    \multicolumn{4}{@{}l}{\emph{\textsc{PTr} (98 q) @ $50$K: BM25 lift (hybrid $-$ dense)}}\\
    Theoria$+$BM25 $-$ Theoria              & $+0.20$ & $[-0.02,\,+0.44]$     & $95$ \\
    Mem0$+$BM25 $-$ Mem0                    & $+0.29$ & $[+0.11,\,+0.48]^{*}$ & $98$ \\
    Simple RAG$+$BM25 $-$ Simple RAG        & $+0.50$ & $[+0.24,\,+0.79]^{*}$ & $98$ \\
    \addlinespace
    \multicolumn{4}{@{}l}{\emph{\textsc{PTr} (98 q) @ $50$K: among the three hybrids}}\\
    Mem0$+$BM25 $-$ Simple RAG$+$BM25       & $+0.03$ & $[-0.16,\,+0.22]$ & $98$ \\
    Mem0$+$BM25 $-$ Theoria$+$BM25          & $+0.06$ & $[-0.20,\,+0.34]$ & $97$ \\
    Simple RAG$+$BM25 $-$ Theoria$+$BM25    & $+0.01$ & $[-0.21,\,+0.24]$ & $97$ \\
    \bottomrule
  \end{tabular}
\end{table}

The conclusions agree with the claims we make from the point estimates.
First, the \textsc{Paim} top cluster is a genuine statistical tie:
the Simple RAG / Theoria / Mem0 differences at $30$K all have intervals
that include $0$, whereas Simple RAG's margins over Hindsight,
Cognee, and the KG-only Graphiti are all significant. Thus, our claim that
bonuses from architecture vanish under budget control is about the top cluster, not about
the weak systems, which really are beaten. 

Second, the BM25 lift on \textsc{PTr} is significant for Mem0 ($+0.29$) and Simple RAG ($+0.50$)
but only a positive {trend} for Theoria ($+0.20$, interval marginally including $0$).
Third, the three hybrids are statistically indistinguishable at $50$K (all pairwise intervals
include $0$ and sit within $\pm0.35$), which is why we do not read the $0.03$-point ordering
among them as a real ranking. The bootstrap script is released in the artifact.

% =============================================================================
\subsection{Judge calibration: multi-judge LLM and human ranking}
\label{sec:judge}
% =============================================================================

The scores in our benchmark come from an LLM judge, so the reliability of
that judge is itself part of the contribution. We therefore compare four LLM
judges and add blinded human pairwise comparisons. As a result, we find that
both system-level ranks and per-question scores correlate well across judges.
There is non-trivial cell-level disagreement on close calls, so single-judge
rankings are reliable for large score gaps but unstable for near-ties, which
is exactly what one would expect from a $10$-point scale.

\begin{figure}[!t]
  \centering\setlength{\tabcolsep}{2pt}
  \begin{tabular*}{\linewidth}{P{.48\linewidth}P{.48\linewidth}}
    \multicolumn{2}{c}{\includegraphics[width=\linewidth]{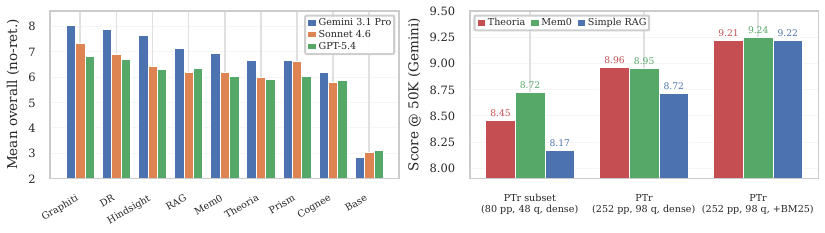}} \\[-2pt]
    \footnotesize \emph{(a)} \textsc{Paim} multi-judge (shared 6q subset) &
    \footnotesize \emph{(b)} \textsc{PTr} corpus $+$ modality at $B\!=\!50$K \\
  \end{tabular*}
  \caption{\emph{(a)}~\textsc{Paim} mean overall by system across three judges on
  the shared 6-question subset, so absolute levels are directly comparable; the
  bar heights track each other closely and rank Spearman is $\rho\in[0.90,0.97]$.
  \emph{(b)}~\textsc{PTr} corpus$+$modality progression at a $50$K retrieval
  budget: the 80-paper / 48-question size-matched subset (dense) $\to$
  \textsc{PTr} dense ($252$ papers, $98$ questions) $\to$ \textsc{PTr} hybrid
  (the same, $+$ BM25). The full-corpus step expands both papers and questions
  over the subset.}
  \label{fig:calibration}
\end{figure}

\textbf{\textsc{Paim} multi-judge subset.} Gemini judged the full
$66$-question $\times$ $9$-protocol \textsc{Paim} grid ($594$ cells); Sonnet~4.6
and GPT-5.4 re-judged a stratified 6-question subset (PQ5, PQ7, PQ13, PQ24,
PQ50, PQ58) across all 9 protocols ($53$ cells per judge, owing to a missing
Graphiti cell on GPT-5.4). Computed on this shared subset, system-rank Spearman
correlations are high: Gemini--Sonnet $\rho=0.90$, Gemini--GPT-5.4 $\rho=0.97$,
Sonnet--GPT-5.4 $\rho=0.93$ (Fig.~\ref{fig:calibration}a). Per-cell agreement
within $1$ point is $90.6\%$ for Sonnet vs.\ GPT-5.4, $69.8\%$ for Gemini vs.\
GPT-5.4, and $57.4\%$ for Gemini vs.\ Sonnet; Gemini is systematically more
generous in absolute level, but the induced ranking is essentially the same.

\begin{figure}[!t]
  \centering
  \begin{tabular*}{\linewidth}{@{\extracolsep{\fill}}cc@{}}
    \multicolumn{2}{c}{\includegraphics[width=\linewidth]{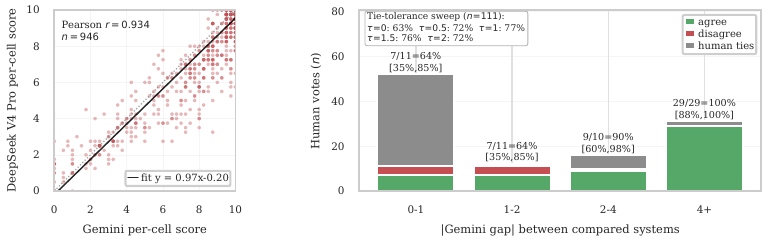}} \\[-2pt]
    \footnotesize \emph{(a)} \textsc{PTr} cross-judge: DeepSeek V4 Pro vs.\ Gemini ($n\!=\!946$) &
    \footnotesize \emph{(b)} Human SxS vs.\ Gemini ($n\!=\!112$) \\
  \end{tabular*}
  \caption{Judge calibration. \emph{(a)}~Per-cell scores, DeepSeek~V4~Pro ($y$)
  vs.\ Gemini~3.1~Pro ($x$) on the \textsc{PTr} 50-question subset ($7$ systems
  $\times$ $50$ questions $\times$ $3$ budgets); Pearson $r=0.93$, $n=946$, best
  fit $y=0.97x-0.20$. \emph{(b)}~Human SxS vs.\ Gemini, $112$ blinded votes.
  Stacked bars count votes per $|\text{Gemini gap}|$ bucket, split into
  ``Gemini agrees with the human winner'', ``Gemini disagrees'', and ``human
  tie''; annotations are agreement rate among decisive humans with Wilson $95\%$
  CIs. The inset reports overall agreement at five tie-tolerance thresholds
  $\tau$; agreement peaks at $\tau\!=\!1$ ($77\%$), confirming that a
  $\sim$$1$-point gap is the effective resolution of the judge.}
  \label{fig:judge_calib}
\end{figure}

\textbf{\textsc{PTr} cross-judge experiment.} We used a 50-question subset of
\textsc{PTr} ($7$ systems $\times$ $50$ questions $\times$ $3$ budgets, $1{,}050$
design cells) to test \emph{DeepSeek~V4~Pro} as a cross-judge against
\emph{Gemini~3.1~Pro} (total cost \$11). After excluding cells where either
judge failed to produce a valid judgment (mostly DeepSeek timeouts on the
base-model $50$K column and a handful of memory-system answers), the paired
comparison contains $n=946$ system/question/budget combinations. Per-cell
Pearson score correlation is $r=0.93$ (Fig.~\ref{fig:judge_calib}a); the
best-fit slope is $0.97$ and the intercept $-0.20$, so DeepSeek is slightly more
critical, and the critical bias concentrates at L3 (the DeepSeek-minus-Gemini
gap is $-0.30$ at L1, $-0.34$ at L2, and $-0.74$ at L3). The two judges disagree
about {which} hybrid is in the first place---under DeepSeek it is Theoria$+$BM25
($9.14$), under Gemini it is Simple~RAG$+$BM25 ($9.53$)---but under both judges
the three hybrids fall within $\sim$$0.3$ of each other and beat their dense
counterparts. We therefore do not treat the top ordering within the hybrid
cluster as meaningful, and our headline result that hybrids $>$ dense with a
three-way tie at the top holds under both judges. Per-system numbers are shown
in Table~\ref{tab:deepseek}.

\begin{table}[!t]
  \centering
  \caption{Per-system mean overall on the 50-question \textsc{PTr} expansion
  subset at $50$K budget under Gemini~3.1~Pro vs.\ DeepSeek~V4~Pro, plus the
  average $\Delta$ across all three budgets. Per-cell Pearson $r=0.93$ over
  $n=946$ valid paired cells; DeepSeek is uniformly $\sim$$0.4$--$0.6$ more
  critical except on the (near-floor) base model.}
  \label{tab:deepseek}
  \footnotesize
  \begin{tabular}{@{}lrrrr@{}}
    \toprule
    \textbf{System} & \textbf{Gemini @ 50K} & \textbf{DeepSeek @ 50K} & \textbf{$\Delta$ at 50K} & \textbf{Avg.\ $\Delta$ (all)} \\
    \midrule
    Theoria             & $9.14$ & $8.61$ & $-0.53$ & $-0.58$ \\
    Theoria + BM25      & $9.48$ & $9.14$ & $-0.34$ & $-0.49$ \\
    Mem0 (dense)        & $9.39$ & $9.11$ & $-0.28$ & $-0.38$ \\
    Mem0 + BM25         & $9.41$ & $8.91$ & $-0.49$ & $-0.40$ \\
    Simple RAG          & $9.32$ & $8.87$ & $-0.44$ & $-0.44$ \\
    Simple RAG + BM25   & $9.53$ & $8.95$ & $-0.58$ & $-0.55$ \\
    Base model          & $1.79$ & $2.18$ & $+0.39$ & $+0.30$ \\
    \bottomrule
  \end{tabular}
\end{table}

\textbf{Blinded human side-by-side study.} Human assessors, blinded to system
identity and to Gemini's scores, voted on $112$ randomly sampled A/B answer
pairs, choosing ``A'', ``B'', or ``tie'' (system identities and Gemini scores
were revealed only \emph{after} each vote, so they could not influence it).
Fig.~\ref{fig:judge_calib}b shows that agreement scales cleanly with the Gemini
gap: among decisive human votes, the judge agrees with the human winner $11/22$
($50\%$) when the gap is $0$--$2$ points, $9/10$ ($90\%$) at $2$--$4$ points,
and $29/29$ ($100\%$) at $4{+}$ points. Allowing a tie tolerance recovers most
of the disagreement: declaring a Gemini ``tie'' whenever $|\text{gap}| \leq
\tau$ lifts overall agreement from $63\%$ at $\tau=0$ to $77\%$ at $\tau=1$, and
the rate stays $\geq 72\%$ for $\tau\in[0.5,2]$. In short, a $1$-point Gemini gap
on a $10$-point scale should not be read as a real ranking difference, while a
$\geq 2$-point gap nearly always agrees with the human verdict. The full
per-system win/loss/tie counts (with Wilson $95\%$ CIs) are released in the artifact.
% \texttt{sxs\_human\_votes.jsonl}, together with the vote schema and the script
% that regenerates Fig.~\ref{fig:judge_calib}.

\textbf{Recommendation.} On scientific QA at this scale, no single judge should
be treated as the ground truth, but with a sensible tie tolerance Gemini agrees with
a human rater on $77\%$ of blinded pairwise comparisons and correlates strongly
with DeepSeek~V4~Pro, Sonnet~4.6, and GPT-5.4. We therefore recommend that
scientific-memory benchmarks:
\begin{enumerate}[label=(\roman*)]
\item report per-cell agreement with a second frontier judge on a stratified
  subset, plus per-question Pearson/Spearman;
\item avoid claiming system-rank differences within $1$--$2$ points on a
  $10$-point scale;
\item validate cell-level reliability with at least a small blinded human SxS
  study.
\end{enumerate}

% =============================================================================
\section{Qualitative analysis}
\label{sec:qualitative}
% =============================================================================

Aggregate scores are important but cannot tell us \emph{why} systems differ. 
We first follow the two questions introduced in Section~\ref{sec:bench} through the systems' answers
and then discuss how the {same} system's answer to the {same} question changes as retrieval budget grows.
All excerpts and scores are from the released run artifacts (Gemini judge, audited rubrics);
several more examples, one per tier per corpus, are given in Appendix~\ref{app:examples}.

\subsection{The same question across systems}
\label{sec:qual_systems}

\textbf{PQ40} (\textsc{Paim}, L2; enumeration $+$ cross-document). The gold
answer is the four MemoryAgentBench competencies and that \emph{selective
forgetting} is the most neglected (Section~\ref{sec:bench}; sources
\cite{hu2025memoryagentbench}, \cite{maharana2024locomo},
\cite{wu2025longmemeval}). The spread is driven entirely by whether a
system's retrieval surfaces the canonical four-item list.

\begin{center}
\footnotesize
\begin{tabular}{@{}>{\raggedright\arraybackslash}p{0.18\linewidth}@{\hspace{4pt}}c@{\hspace{6pt}}>{\raggedright\arraybackslash}p{0.66\linewidth}@{}}
  \toprule
  \textbf{System} & \textbf{Score} & \textbf{Answer (excerpt)} \\
  \midrule
  Simple RAG    & $10.0$           & ``MemoryAgentBench evaluates: (1)~Accurate Retrieval, (2)~Test-Time Learning, (3)~Long-Range Understanding, (4)~Selective Forgetting. Selective Forgetting is the most often neglected by LoCoMo and LongMemEval.'' \\
  \addlinespace[1pt]
  Theoria       & $\phantom{0}9.0$ & ``Four competencies: Accurate Retrieval (AR), Test-Time Learning (TTL), Long-Range Understanding (LRU), Selective Forgetting (SF). Prior benchmarks such as LoCoMo and LongMemEval most often neglect Long-Range Understanding (LRU) and Selective Forgetting (SF)\ldots'' [partially correct -- adds LRU, which is not in the gold.] \\
  \addlinespace[1pt]
  Hindsight     & $\phantom{0}6.0$ & ``Four core competencies: Accurate Retrieval (AR), Test-Time Learning (TTL), Long-Range Understanding (LRU), Selective Forgetting (SF). Prior benchmarks such as LoCoMo and LongMemEval primarily focus on aspects like long-term interaction\ldots'' \\
  \addlinespace[1pt]
  Graphiti (KG) & $\phantom{0}0.5$ & ``MemoryAgentBench evaluates memory agents focusing on their memory capabilities but does not evaluate agentic actions or environment feedback\ldots The exact four core competencies are not explicitly listed in the evidence, we can infer\ldots'' \\
  \bottomrule
\end{tabular}
\end{center}

\noindent Simple RAG pulls the source enumeration cell verbatim; Theoria and
Hindsight retrieve the list but append a non-gold item (LRU) and lose a point or
more; the KG-only Graphiti, whose edges/nodes never surfaced the list, can only
hedge ($0.5$). In this case, it is mostly not about reasoning but simply about whether the
exact list reached the synthesizer.

\textbf{TX17} (\textsc{PTr}, L2; quantitative $+$ cross-document). Here the gold
answer spans five papers and hinges on rare surface forms (``Kimi Linear
$6\times$ at $1$M''); sources \cite{yan2025fsa}, \cite{mu2025sals},
\cite{lingteam2025ringflashlinear}, \cite{hu2025ultralong},
\cite{kimiteam2025kimilinear}. This is where retrieval \emph{modality} becomes
important.

\begin{center}
\footnotesize
\begin{tabular}{@{}>{\raggedright\arraybackslash}p{0.20\linewidth}@{\hspace{4pt}}c@{\hspace{6pt}}>{\raggedright\arraybackslash}p{0.64\linewidth}@{}}
  \toprule
  \textbf{System} & \textbf{Score} & \textbf{Answer (excerpt)} \\
  \midrule
  Theoria             & $10.0$           & ``FSA: improvements primarily at $\geq 64$K, up to $3.5\times$ kernel speedup; SALS: useful from $4$K with $6.4\times$ KV compression; Kimi Linear: $6\times$ decoding throughput at $1$M context (most aggressive)\ldots Ring-flash-linear-2.0: linear-attention pronounced beyond $8$K\ldots'' \\
  \addlinespace[1pt]
  Theoria $+$ BM25    & $10.0$           & ``FSA: $\sim$$8$K--$16$K crossover; SALS: efficient from $4$K, scales to RULER-$128$K; Kimi Linear: $6\times$ decoding at $1$M (most aggressive); Ring-flash-linear-2.0: pronounced beyond $8$K; HSA-UltraLong: $>90\%$ retrieval at $16$M tokens.'' \\
  \addlinespace[1pt]
  Simple RAG $+$ BM25 & $10.0$           & ``Comprehensive comparison across FSA, SALS, Kimi Linear, Ring-flash-linear-2.0, HSA-UltraLong\ldots Kimi Linear gives the most aggressive throughput claim at $1$M$+$ tokens with $6\times$ decoding throughput.'' \\
  \addlinespace[1pt]
  Simple RAG          & $\phantom{0}5.0$ & ``FSA becomes beneficial beyond $\sim$$8$K--$16$K tokens. At shorter sequences\ldots'' [retrieval pulled relevant FSA paragraphs but missed Kimi Linear's $1$M-context claim, giving an incomplete answer.] \\
  \bottomrule
\end{tabular}
\end{center}

\noindent Dense Simple RAG retrieves the FSA discussion but never the rare
``Kimi Linear $6\times$'' phrase, and scores $5.0$; adding BM25 surfaces that
exact string and lifts the same system to $10.0$. This is a representative
example of the \textsc{PTr} modality story of Section~\ref{sec:ptr_results}.

\subsection{The same system across budgets}
\label{sec:qual_budget}

A larger token budget usually helps on average, but per question the effect is sharper and
not always monotone. We show two representative cases.

\textbf{Budget unlocks a correct answer (Simple RAG, PQ9).}

\begin{quote}
\noindent\textbf{PQ9} (\textsc{Paim}, L2; mechanistic $+$ quantitative).\\
\textbf{Q:} ``Why does RAPTOR's collapsed-tree retrieval outperform tree
traversal, and by how much? What does the paper identify as the mechanism, and
how does the benefit change with model strength?''\\
\textbf{Gold:} collapsed-tree retrieval outperforms tree traversal by
$2$--$4.5$ points; flattening all levels lets every leaf and summary compete by
cosine similarity, avoiding the routing errors that top-down traversal
accumulates; weaker reader models benefit more from the tree structure.\\
\textbf{Sources:} \cite{sarthi2024raptor}
\end{quote}

\noindent At $10$K the top-$k$ surfaces a passage \emph{describing} tree
traversal and the synthesizer confidently asserts the wrong winner; more budget
surfaces the actual comparison, and the answer becomes correct and stays correct.

\begin{center}
\footnotesize
\begin{tabular}{@{}c@{\hspace{6pt}}c@{\hspace{6pt}}>{\raggedright\arraybackslash}p{0.70\linewidth}@{}}
  \toprule
  \textbf{Budget} & \textbf{Score} & \textbf{Answer (excerpt) and judge verdict} \\
  \midrule
  $10$K & $\phantom{0}0.0$  & ``\ldots The paper reports \textbf{tree traversal} as the consistently better-performing retrieval strategy\ldots'' \emph{Judge:} ``completely fails by claiming tree traversal is the winner instead of collapsed-tree retrieval, \ldots\ invents mechanisms to support this incorrect claim.'' \\
  \addlinespace[2pt]
  $30$K & $10.0$ & ``\ldots The \textbf{collapsed tree retrieval} strategy consistently outperforms the tree traversal method across experiments\ldots supported by results on QASPER (Figure~3)\ldots'' \emph{Judge:} ``perfectly identifies collapsed-tree retrieval as the winner and accurately explains the mechanism.'' \\
  \addlinespace[2pt]
  $50$K & $10.0$ & ``\ldots the \textbf{collapsed tree retrieval} strategy consistently outperforms the tree traversal method across experiments\ldots'' \emph{Judge:} ``perfectly captures the required facts\ldots without triggering any penalties.'' \\
  \bottomrule
\end{tabular}
\end{center}

\noindent Under-budgeted retrieval here does not merely {truncate} the
answer; it shows a plausible-but-wrong passage and yields a confident
$0.0$. 
% The architecture is unchanged---only the number of retrieved characters
% moved.

\textbf{Higher budget introduces a distractor (Theoria, PQ44).}

\begin{quote}
\noindent\textbf{PQ44} (\textsc{Paim}, L2; mechanistic $+$ cross-document).\\
\textbf{Q:} ``HaluMem decomposes memory hallucinations into three stage-specific
error categories. Name each stage and explain why extract-and-update systems
like Mem0 and Memory-R1 are structurally more vulnerable than verbatim-storage
systems like MemMachine.''\\
\textbf{Gold:} the three stages are \emph{extraction}, \emph{updating}, and
\emph{memory-QA}; extract-and-update systems write hallucinated facts permanently
(and later retrieve them as if true), whereas verbatim-storage systems such as
MemMachine extract only at read time, so errors do not compound in the store.\\
\textbf{Sources:} \cite{halumem}, \cite{mem0}, \cite{yan2025memoryr1}, \cite{wang2026memmachine}
\end{quote}

\noindent At $10$--$30$K Theoria answers correctly; at $50$K its
community-routed retrieval pulls in a {neighboring} paper (EverMemOS), and
the synthesizer conflates the two, substituting EverMemOS's memory stages for
HaluMem's.

\begin{center}
\footnotesize
\begin{tabular}{@{}c@{\hspace{6pt}}c@{\hspace{6pt}}>{\raggedright\arraybackslash}p{0.70\linewidth}@{}}
  \toprule
  \textbf{Budget} & \textbf{Score} & \textbf{Answer (excerpt) and judge verdict} \\
  \midrule
  $10$K & $\phantom{0}9.25$ & ``\ldots three distinct, stage-specific error categories\ldots Extraction\ldots Update\ldots'' \emph{Judge:} ``accurately identifies the three stages and the structural vulnerabilities, though the retrieved sources completely missed the HaluMem paper.'' \\
  \addlinespace[2pt]
  $30$K & $\phantom{0}9.5$  & ``\ldots three distinct, stage-specific error categories: (1)~Extraction Errors, (2)~Update Errors, (3)~[Memory-QA]\ldots'' \emph{Judge:} ``accurately identifies the three stages\ldots\ slightly misses the explicit nuance that hallucinated facts become permanently written.'' \\
  \addlinespace[2pt]
  $50$K & $\phantom{0}4.5$  & ``\ldots three distinct stages\ldots (1)~\textbf{Episodic Trace Formation}\ldots'' \emph{Judge:} ``failed to retrieve the HaluMem paper and incorrectly substituted EverMemOS's memory stages for HaluMem's, missing the core argument.'' \\
  \bottomrule
\end{tabular}
\end{center}

\noindent This is the candidate-pool dilution of Section~\ref{sec:ptr_results}
in miniature: the extra $\sim$$20$K characters at $50$K were not neutral filler
but a semantically adjacent distractor that displaced the correct source. This
example shows that ``more retrieval'' is not uniformly better, and that the
budget--quality curve can be non-monotone for an individual question even when it
rises on average (Appendix~\ref{app:examples} gives a third, oscillating case).

% =============================================================================
\section{Discussion and limitations}
\label{sec:discussion}
% =============================================================================

\subsection{Conclusions from the evaluation study}

Our experiments show that there is no definitive architecture-only leaderboard
for scientific agentic memory. On \textsc{Paim}, native results are dominated by
raw-context volume (Graphiti's $8.04$ native overall comes with $2.55$M chars
per query; its KG-only ablation drops to $5.27$ at $30$K), and dense
budget-targeted retrieval leaves Simple RAG, Theoria, and Mem0 within $0.3$ of
one another. On \textsc{PTr}, which has clearer named entities, architecture
lineages, and quantitative technical claims, structured and extracted memories
(Theoria, Mem0) edge ahead of Simple RAG by $0.2$--$0.3$ under dense retrieval at
the same budget. Adding BM25 to the three evaluated systems on \textsc{PTr}
produces the largest measured improvement and makes their overall scores
converge within $0.03$ at $50$K. Budget control is thus \emph{necessary but not
sufficient}: modality, corpus structure, and representation density all matter,
and benchmarks should report all of them.

\textbf{What this says about Theoria.} Theoria does not win the overall hybrid
leaderboard, but this is not a negative result. Under dense retrieval it is
competitive with Simple RAG on \textsc{Paim} and stronger on \textsc{PTr},
especially at high budget; with BM25 it joins the same top cluster as Mem0 and
Simple RAG. This suggests that theory/community structure can improve budgeted
evidence selection in some corpora, but that retrieval modality is a larger
lever for the single-shot QA protocol studied here. The parts of Theoria most
unlike RAG---\texttt{/theories} as a cold-start literature map and
\texttt{/observe} as an incremental theory-update loop---remain untested by
one-shot QA and motivate future sequential research-agent benchmarks.

\textbf{Compression versus evidence preservation.} The budgeted results refine
the usual intuition that extraction loses information. Mem0's atomic facts are
{competitive} once retrieval is matched, so aggressive extraction is not
inherently bad; what fails is discarding the answer-critical details
(quantities, conditions, table values) that L2/L3 questions require. The KG-only
Graphiti row, which keeps structure but throws away raw episode text, is the
clearest illustration: it is competitive on some L3 synthesis but collapses on
L1 factual recall, because the exact numbers are in the episode bodies that
it no longer returns.

\textbf{How much can routing help?} The tier-shaped specialization in
Tables~\ref{tab:paim}--\ref{tab:ptr} (Theoria$+$BM25 wins L1, Mem0$+$BM25 wins
L3) suggests a question-conditioned ensemble. Because we already have every
system's answer and judge score on every question, we can measure the ceiling
{without any new inference}, as a pure re-aggregation of
existing scores. 

On \textsc{PTr} (98 questions) at $50$K, an \emph{oracle} router that
picks the best-scoring of the three hybrids per question reaches $9.68$, a
$+0.40$ gain over the best single hybrid; over all six dense and hybrid systems
the oracle reaches $9.73$ ($+0.46$). On \textsc{Paim} at $50$K the dense
top-cluster oracle is $8.39$ vs.\ $7.61$ for the best single system ($+0.77$).
The headroom is therefore real and significant.

The catch is that a tier-level router that knows only the difficulty
tier does not capture it: a leave-one-question-out, tier-conditioned router
scores $9.22$ on \textsc{PTr} and $7.12$ on \textsc{Paim}, which is no better than
(even slightly below) simply always using the best single system.
The oracle's advantage comes from {per-question}
variation that the tier label (and, in additional checks, the question type
label) does not predict. So in principle, a router conditioned on the question
could add another $\sim$$0.4$--$0.8$ points, but
realizing that gain is an open problem that requires a per-question routing
signal, most probably a separate LLM run.

\textbf{Recommendations.} Based on our results we make the following
recommendations for future scientific-memory benchmarks.
\begin{enumerate}
\item Report mean retrieved characters per query alongside scores.
\item Report retrieval modality (dense, lexical, hybrid).
\item Always include Simple~RAG \emph{and} Simple~RAG$+$BM25 as null
  hypotheses; dense-only RAG alone is too weak a null.
\item Run a budget-targeted track in addition to the native track, and report
  achieved characters rather than nominal budgets.
\item Audit rubrics against full paper text before reporting rankings (we caught
  many errors in the original data).
\item Cross-validate judges and report Spearman $\rho$ on a shared question
  subset rather than treating any single LLM as ground truth; calibrate
  cell-level reliability with at least a small human SxS study.
\item Document granularity contracts and any deviations from them.
\end{enumerate}

% =============================================================================
\subsection{Limitations}
\label{sec:limits}
% =============================================================================

We also note the following limitations of our study.
\begin{enumerate}
\item \emph{Modest question count} ($66$ / $98$); within-cluster differences
  smaller than $\sim$$0.3$ should be treated as noise.
\item \emph{Sparse \textsc{PTr} coverage}: the $98$ questions reach only
  $\sim$$48\%$ of the $252$-paper corpus, leaving $132$ papers uncovered.
\item \emph{Approximate budget control}: three systems do not honor a clean
  character cap (Hindsight ignores \texttt{n}; Cognee's chunks are coarse;
  Graphiti's \texttt{top\_k} is per-modality), so reported volumes are close but
  not exactly equal across systems.
\item \emph{Synthesis model fixed} (gpt-4.1-mini, $T{=}0$); a comparison across
  synthesis models is interesting future work.
\item \emph{Single run per (system, budget)} due to cost; smaller deltas
  ($\pm 0.05$) would benefit from multi-seed reruns.
\item \emph{Human SxS sample is relatively small}; although the agreement curve
  is convincing as far as it goes, broader human evaluation would strengthen it.
\item \emph{Public-paper contamination}: because the corpora are public arXiv
  papers, the no-retrieval base model is not a pure zero-information baseline,
  and markdown conversion can introduce noise around tables and equations.
\end{enumerate}
These limitations are, to a large degree, precisely why memory-system evaluation
needs the kind of explicit, budget-aware protocol we propose.

% =============================================================================
\subsection{Lessons for memory evaluation infrastructure}
\label{sec:doctor}
% =============================================================================

Several integration patterns silently produced misleading numbers during our
experiments, and they motivate a pre-run ``doctoring'' phase that probes each
system with a known query before any evaluation begins. We report them
because they are the kind of failures that an uncontrolled leaderboard
might absorb without warning.
\begin{enumerate}
\item A Qdrant collection symlink on \textsc{Paim} pointed to an empty directory
  after a working-directory change; \texttt{mem.search()} then returned $0$
  characters on most queries with no exception raised.
\item A ``\texttt{fastembed} not installed'' warning at ingest time silently
  disabled BM25 hybrid retrieval for an entire phase; only re-installation
  revealed that Mem0 had been running dense-only the whole time.
\item A Qdrant local-mode SDK crashed with an out-of-memory error at $38{,}330$
  points (the SDK warns it is ``not recommended above $20{,}000$''); the failure
  surfaced only when a later query died at load time.
\end{enumerate}

We recommend a six-line checklist before any memory-system run:
\begin{enumerate}[label=(\arabic*)]
\item a known good query returns nonzero retrieved context;
\item the retrieved character count matches the expected budget within a
  tolerance;
\item lexical and dense modalities are actually enabled;
\item the stored-unit count matches the ingest logs;
\item the source-document count matches the corpus manifest;
\item the \texttt{top\_k} knob actually changes output size on a probe query.
\end{enumerate}

% =============================================================================
\section{Conclusion}
\label{sec:conclusion}
% =============================================================================

We argue that scientific memory should be evaluated as budgeted, modality-aware
\emph{context restoration} rather than as an unconstrained architecture
leaderboard, and we built the artifacts to do so. We release
\begin{enumerate}[label=(\roman*)]
\item \emph{two full-text scientific-memory benchmarks}, \textsc{Paim}
($81$~papers, $66$ audited questions) on AI agent memory and \textsc{PTr}
($252$~papers, $98$ questions) on modern transformer literature;
\item \emph{a context-restoration evaluation harness} that explicitly varies
retrieval budget and modality while holding the synthesis model fixed, applied
to eight memory systems plus a no-retrieval baseline, with BM25 hybrid variants
for three of them; 
\item \emph{Theoria}, a three-layer
evidence~$+$~community~$+$~theory memory system, also evaluated as a
participant; and 
\item \emph{an LLM-as-judge protocol} verified against human
votes and cross-validated across frontier LLMs.
\end{enumerate}
All of these artifacts are public: the harness, adapters, raw outputs,
judgments, and analysis scripts at
\url{https://gitlab.com/quantellence/research/scientific-recall-bench}, and the
\textsc{Paim} and \textsc{PTr} corpora with their audited rubrics at
\url{https://huggingface.co/datasets/quantellence/srb-data}.

We found that native leaderboards conflate architecture with retrieved evidence
volume: Graphiti's lead on \textsc{Paim} comes with $2.55$M characters per
query, and once retrieval is constrained the lead over Simple RAG vanishes on
\textsc{Paim} and is small ($0.2$--$0.3$ points) on \textsc{PTr}. Budget control
is necessary but not sufficient: adding BM25 hybrid retrieval to the systems
where it can be wired up cleanly is the single largest intervention we measured,
and the three resulting hybrids on \textsc{PTr} tie within $0.03$ points at $50$K
characters despite ingest costs that span four orders of magnitude. Multi-judge
and human SxS calibration further show that a Gemini score gap below $\sim$$1$
point is below the noise floor of the protocol; with a $\tau{=}1$ tolerance, the
LLM judge agrees with a blinded human rater on $77\%$ of $112$ pairs, and the
agreement curve scales cleanly with the gap.

We note several directions for future work. First, we considered only one-shot, single-question
retrieval; the parts of \emph{Theoria} most unlike chunk RAG, namely the cold-start
\texttt{GET /theories} endpoint and the iterative \texttt{POST /observe}
loop, have not been evaluated at all. The tier specialization we observed
(Theoria$+$BM25 wins L1, Mem0$+$BM25 wins L3) points to an ensemble opportunity:
a per-question oracle over our systems would add $0.4$--$0.8$ points
(Section~\ref{sec:discussion}), but a tier-only router does not realize it, so a
per-question routing signal is the open problem. \textsc{PTr} questions reach
only $\sim$$48\%$ of the corpus, so expanding question coverage, as well as adding
scientific domains beyond AI memory and Transformers, would test how far the
corpus structure / retrieval modality story generalizes. Finally, and most
consequentially in our view, the field needs a \emph{sequential}
research-agent benchmark rather than one-shot QA. We see scientific memory as a
research-agent infrastructure problem rather than a static leaderboard, and we
hope this protocol and its artifacts make that infrastructure progressively more
measurable.

% =============================================================================
% \begin{ack}
% The work of Sergey Nikolenko was supported by the Ministry of Economic Development of the Russian Federation (Agreement No. 139-15-2025-007, dated April 16, 2025; ID: 000000C313925P3O0002).
% \end{ack}

\bibliographystyle{plainnat}
\bibliography{references}

%%%%%%%%%%%%%%%%%%%%%%%%%%%%%%%%%%%%%%%%%%%%%%%%%%%%%%%%%%%%%%%%%%%%%%%%%%%%%%%%%
\appendix

\section{Qualitative answer examples}
\label{app:examples}

This appendix collects additional per-tier examples beyond the cases considered in
Section~\ref{sec:qualitative} (PQ40, TX17, and the two budget-sensitivity cases).
Each is chosen for cross-system disagreement (high standard deviation of judge
scores across systems); we show the question, a one-line gold-answer summary, and
three to four representative system answers, each with the Gemini judge's overall
score for that cell. Answers are abbreviated to the first $\sim$$3$ sentences for
readability; the full answers are in the released artifact
(\texttt{synthesized.jsonl}). The final subsection adds a third
budget-sensitivity example to complement Section~\ref{sec:qual_budget}.

\subsection*{PQ4 -- saturation with integration failures}

\begin{quote}
\noindent\textbf{PQ4} (\textsc{Paim}, L1; factual $+$ quantitative).\\
\textbf{Q:} ``What Judge score does MAGMA report on LoCoMo, and by what margin
does it beat prior systems according to the paper?''\\
\textbf{Gold:} MAGMA achieves an LLM-as-a-Judge score of $0.700$ on LoCoMo,
beating Full Context ($0.481$), A-MEM ($0.580$), MemoryOS ($0.553$), and Nemori
($0.590$) by $18.6\%$--$45.5\%$ relative.\\
\textbf{Sources:} \cite{jiang2026magma}, \cite{maharana2024locomo}
\end{quote}

\begin{center}
\footnotesize
\begin{tabular}{@{}>{\raggedright\arraybackslash}p{0.18\linewidth}@{\hspace{4pt}}c@{\hspace{6pt}}>{\raggedright\arraybackslash}p{0.66\linewidth}@{}}
  \toprule
  \textbf{System} & \textbf{Score} & \textbf{Answer (excerpt)} \\
  \midrule
  Theoria       & $10.0$ & ``MAGMA reports an overall LLM-as-a-Judge score of $0.700$ on LoCoMo. This score significantly outperforms prior systems, beating Full Context ($0.481$) by $45.5\%$, A-MEM ($0.580$) by $18.6\%$, MemoryOS ($0.553$) by $\sim$$26.6\%$, and Nemori ($0.590$) by $\sim$$18.6\%$.'' \\
  \addlinespace[1pt]
  Simple RAG    & $10.0$ & ``MAGMA reports an overall LLM-as-a-Judge score of $0.7$ on the LoCoMo benchmark, with relative improvements of $18.6\%$--$45.5\%$ over Full Context ($0.481$), A-MEM ($0.58$), MemoryOS ($0.553$) and Nemori ($0.59$).'' \\
  \addlinespace[1pt]
  Mem0          & $10.0$ & ``MAGMA reports an overall LLM-as-a-Judge score of $0.7$ ($70\%$) on LoCoMo, substantially outperforming Full Context, A-MEM, MemoryOS, and Nemori by relative margins of $18.6\%$ to $45.5\%$.'' \\
  \addlinespace[1pt]
  Graphiti (KG) & $\phantom{0}4.0$ & ``The provided evidence does not specify the exact LLM-as-a-Judge score MAGMA achieves on LoCoMo, nor the precise margins. MAGMA is described as a multi-graph agentic memory architecture\ldots'' [the relevant numbers were in the episode body that the KG-only mode does not return.] \\
  \bottomrule
\end{tabular}
\end{center}

\emph{Reading.} The dense top systems all return the same correct numbers; the
spread is driven entirely by systems whose retrieval scope does not surface the
relevant rubric facts.

\subsection*{PQ28 -- multi-paper density argument}

\begin{quote}
\noindent\textbf{PQ28} (\textsc{Paim}, L3; quantitative $+$ conditional $+$ cross-document).\\
\textbf{Q:} ``GraphRAG-Bench argues that graph density predicts performance more
than architectural sophistication. State the specific average-degree and
edges-per-10K-tokens numbers for HippoRAG~2 and MS-GraphRAG, and describe what
this implies for systems whose graphs fall below the density threshold.''\\
\textbf{Gold:} HippoRAG~2 avg degree $8.75$ (novel) / $13.31$ (medical),
$\sim$$2{,}310$ edges per 10K tokens (novel); MS-GraphRAG avg degree $1.48$ /
$1.82$, $\sim$$273$ edges per 10K tokens. Sparser graphs underperform regardless
of architecture.\\
\textbf{Sources:} \cite{xiang2025graphragbench}, \cite{gutierrez2024hipporag}, \cite{gutierrez2025hipporag2}, \cite{edge2024graphrag}
\end{quote}

\begin{center}
\footnotesize
\begin{tabular}{@{}>{\raggedright\arraybackslash}p{0.18\linewidth}@{\hspace{4pt}}c@{\hspace{6pt}}>{\raggedright\arraybackslash}p{0.66\linewidth}@{}}
  \toprule
  \textbf{System} & \textbf{Score} & \textbf{Answer (excerpt)} \\
  \midrule
  Simple RAG    & $10.0$           & ``HippoRAG~2 average node degree: novel $8.75$, medical $13.31$. MS-GraphRAG: novel $1.48$, medical $1.82$. HippoRAG~2 produces $\sim$$2{,}310$ edges on the novel dataset, vs.\ much sparser graphs from MS-GraphRAG\ldots Systems whose graphs fall below this density threshold consistently underperform.'' \\
  \addlinespace[1pt]
  Cognee        & $10.0$           & ``[GraphRAG-Bench] Average node degree: HippoRAG~2 $8.75$/$13.31$ vs.\ MS-GraphRAG $1.48$/$1.82$; edges per 10K tokens $\sim$$2{,}310$ for HippoRAG~2 (novel) vs.\ $\sim$$273$ for MS-GraphRAG\ldots Systems below the density threshold (e.g.\ MS-GraphRAG) lose answer accuracy proportionally.'' \\
  \addlinespace[1pt]
  Theoria       & $\phantom{0}7.8$ & ``HippoRAG~2: novel $8.75$, medical $13.31$\ldots edges-per-10K-tokens for HippoRAG~2 is not explicitly stated, we can infer from the data that HippoRAG~2 generates significantly denser graphs with $2{,}310$ edges on the novel dataset.'' \\
  \addlinespace[1pt]
  Mem0          & $\phantom{0}6.0$ & ``HippoRAG~2: novel $8.75$, medical $13.31$. Exact average degrees for MS-GraphRAG are not explicitly stated in the excerpts\ldots'' \\
  \addlinespace[1pt]
  Graphiti (KG) & $\phantom{0}2.0$ & ``The provided evidence does not include explicit numerical values\ldots HippoRAG~2 is evaluated on GraphRAG-Bench and shows improved QA performance compared to GraphRAG and other baselines.'' \\
  \bottomrule
\end{tabular}
\end{center}

\emph{Reading.} This synthesis question rewards systems whose retrieval scope
returns multiple paragraphs from a single source: Cognee's coarse $\sim$$19$K
chunks are well matched here (and in fact tie Simple RAG); Theoria and Mem0
surface part of the evidence but miss the second paper's numbers; Graphiti~(KG)
loses everything that lived in the episode body (which native Graphiti would
place in its huge retrieved context).

\subsection*{TX4 -- BM25 closes a small gap}

\begin{quote}
\noindent\textbf{TX4} (\textsc{PTr}, L1; factual $+$ enumeration).\\
\textbf{Q:} ``What two components make up DeepSeek Sparse Attention (DSA) layered
on top of MLA, and what context length does DeepSeek-V3.2 target?''\\
\textbf{Gold:} DSA $=$ lightning indexer $+$ fine-grained (top-$k$) token
selection mechanism, instantiated under MLA; DeepSeek-V3.2 targets $128$K
context (not $1$M).\\
\textbf{Sources:} \cite{deepseekai2025v32}
\end{quote}

\begin{center}
\footnotesize
\begin{tabular}{@{}>{\raggedright\arraybackslash}p{0.20\linewidth}@{\hspace{4pt}}c@{\hspace{6pt}}>{\raggedright\arraybackslash}p{0.64\linewidth}@{}}
  \toprule
  \textbf{System} & \textbf{Score} & \textbf{Answer (excerpt)} \\
  \midrule
  Simple RAG       & $10.0$           & ``DSA primarily consists of two components: (1)~Lightning Indexer, which computes an index score $I_{t,s}$ between query and preceding tokens; (2)~Fine-Grained Token Selection Mechanism\ldots DeepSeek-V3.2 targets $128$K context.'' \\
  \addlinespace[1pt]
  Theoria $+$ BM25 & $10.0$           & ``DSA components: (1)~Lightning Indexer (low-precision FP8 attention indexer); (2)~Fine-Grained Token Selection Mechanism (top-$k$ KV per query token). Layered under MLA. DeepSeek-V3.2 targets a $128$K context length.'' \\
  \addlinespace[1pt]
  Theoria          & $\phantom{0}7.2$ & ``DSA consists of two main components: Lightning Indexer (FP8 small-head attention indexer\ldots) and Fine-Grained Token Selection Mechanism\ldots'' [missing the explicit ``$128$K context'' fact in the cell the judge scored.] \\
  \addlinespace[1pt]
  Mem0             & $\phantom{0}7.2$ & ``DSA: Lightning Indexer $+$ Fine-Grained Token Selection Mechanism. DSA is instantiated on top of masked multi-head attention\ldots DeepSeek-V3.2 targets $128{,}000$ tokens.'' [the ``masked multi-head attention'' phrasing is judged a partial mismatch with the gold ``MLA''.] \\
  \bottomrule
\end{tabular}
\end{center}

\emph{Reading.} BM25 closes a small but real gap by surfacing the exact fact
strings (e.g., the ``$128$K'' value in a context paragraph) that dense retrieval
ranked below the cut.

\subsection*{TX37 -- five trends $\times$ flagship models}

\begin{quote}
\noindent\textbf{TX37} (\textsc{PTr}, L3; synthesis $+$ cross-document).\\
\textbf{Q:} ``Synthesize the five converging trends in the corpus's closing
sections (sparse/linear hybrids, ultra-sparse trillion MoE, test-time compute,
sub-4-bit training and KV-quantization, native multimodal). For each trend, name
two flagship models and cite specific benchmark numbers for at least one trend.''\\
\textbf{Gold:} Sparse/linear: Kimi Linear ($75\%$ KV reduction, $6\times$
decoding at $1$M), DeepSeek-V3.2 (DSA on MLA). Ultra-sparse trillion MoE: Kimi
K2 ($1$T/$32$B), Ring-1T ($1$T/$\sim$$50$B). Test-time compute: TRM ($7$M
params, $45\%$ ARC-AGI-1), Ouro ($2.6$B looped matches $12$B). Sub-4-bit: NVFP4
($12$B$\times$$10$T tokens), KVTC (up to $20\times$ KV compression). Native
multimodal: Emu3.5, Qwen3-VL.\\
\textbf{Sources:} corpus-wide synthesis; cited flagship models include
\cite{kimiteam2025kimilinear} and \cite{deepseekai2025v32}.
\end{quote}

\begin{center}
\footnotesize
\begin{tabular}{@{}>{\raggedright\arraybackslash}p{0.20\linewidth}@{\hspace{4pt}}c@{\hspace{6pt}}>{\raggedright\arraybackslash}p{0.64\linewidth}@{}}
  \toprule
  \textbf{System} & \textbf{Score} & \textbf{Answer (excerpt)} \\
  \midrule
  Mem0 $+$ BM25       & $10.0$           & ``(a)~Sparse/linear hybrids: Kimi Linear ($75\%$ KV reduction, $6\times$ at $1$M), DeepSeek-V3.2 (DSA over MLA); (b)~ultra-sparse trillion MoE: Kimi K2, Ring-1T; (c)~test-time compute: TRM ($7$M, $45\%$ ARC-AGI-1), Ouro ($2.6$B); (d)~sub-4-bit + KV: NVFP4, KVTC; (e)~native multimodal: Emu3.5, Qwen3-VL.'' \\
  \addlinespace[1pt]
  Simple RAG $+$ BM25 & $\phantom{0}5.8$ & ``(a)~Sparse/linear hybrids: hybrid architectures combining sparse or linear attention with full attention\ldots'' [structurally correct but mostly summary prose with few specific model names per trend.] \\
  \addlinespace[1pt]
  Theoria             & $\phantom{0}5.5$ & ``(a)~Sparse/Linear Attention Hybrids: hybrids that combine sparse, linear, and dense attention\ldots'' [similar issue: trend descriptions strong, flagship-model coverage thin.] \\
  \addlinespace[1pt]
  Theoria $+$ BM25    & $\phantom{0}4.0$ & ``Sparse/linear hybrids: hybrid attention architectures combining sparse, linear, and full quadratic attention\ldots'' [BM25 surface match retrieved a different cluster of paragraphs and the synthesis lost specificity.] \\
  \bottomrule
\end{tabular}
\end{center}

\emph{Reading.} A synthesis question at this scope rewards a single retrieval
pass that returns specific numerical claims from many papers; only Mem0$+$BM25
happened to produce that mix here. This supports the routing-ensemble argument
of Section~\ref{sec:ptr_results}: different hybrids win different L3 questions,
and a question-conditioned routing layer would raise the ceiling further.

\subsection*{PQ52 -- sparse-graph failure modes (budget-sensitive)}

\begin{quote}
\noindent\textbf{PQ52} (\textsc{Paim}, L2; conditional $+$ negative $+$ cross-document).\\
\textbf{Q:} ``GraphRAG-Bench identifies that graph-based memory systems fail when
graph density falls below a threshold. Which specific systems in the corpus are
at risk under this criterion? Cite at least one paper (LinearRAG, LightRAG, or
MS-GraphRAG) where density metrics or F1 catastrophes are reported.''\\
\textbf{Gold:} a density threshold near average degree $2.0$; at-risk systems
include MS-GraphRAG (average degree $1.48$) and LightRAG (F1 $=6.6$ catastrophe),
while LinearRAG reports $89.08\%$ recall and HippoRAG~2 is safe (average degree
$8.75$).\\
\textbf{Sources:} \cite{xiang2025graphragbench}, \cite{guo2024lightrag}, \cite{zhuang2025linearrag}, \cite{gutierrez2024hipporag}, \cite{gutierrez2025hipporag2}
\end{quote}

\noindent Simple RAG's score is \emph{non-monotone} in the budget---it dips
sharply at $30$K before recovering at $50$K---complementing the monotone PQ9 and
the over-budget collapse of PQ44 in Section~\ref{sec:qual_budget}.

\begin{center}
\footnotesize
\begin{tabular}{@{}c@{\hspace{6pt}}c@{\hspace{6pt}}>{\raggedright\arraybackslash}p{0.70\linewidth}@{}}
  \toprule
  \textbf{Budget} & \textbf{Score} & \textbf{Answer (excerpt) and judge verdict} \\
  \midrule
  $10$K & $6.0$  & ``\ldots graph-based memory systems fail when graph density falls below a certain threshold\ldots'' \emph{Judge:} ``correctly identifies the systems at risk and the general relationship, but fails to provide the specific density metrics or F1 scores due to poor retrieval.'' \\
  \addlinespace[2pt]
  $30$K & $2.5$  & ``\ldots graph density---measured by metrics such as average degree---governs performance\ldots'' \emph{Judge:} ``triggers a direct penalty by claiming LightRAG is fundamentally sound and avoids pitfalls, completely misses the F1 catastrophe, and hallucinates LinearRAG's failure.'' \\
  \addlinespace[2pt]
  $50$K & $8.25$ & ``\ldots low graph density leads to fragmented or sparse evidence\ldots'' \emph{Judge:} ``correctly identifies MS-GraphRAG and LightRAG as at-risk systems and cites LinearRAG regarding F1 catastrophes, fulfilling the core requirements, though it misses the specific gold metrics.'' \\
  \bottomrule
\end{tabular}
\end{center}

\emph{Reading.} The $30$K retrieval happened to surface a passage that led the
synthesizer to a penalized claim (LightRAG ``fundamentally sound''); the larger
$50$K pool diluted that distractor and recovered most of the answer. As in PQ44,
budget and answer quality are not monotonically related for an individual
question, even though the per-tier averages rise smoothly with budget.

\end{document}